%% file: main.tex
\DocumentMetadata{}
\documentclass[manuscript, nonacm]{acmart}
\input{variables}
\newcommand\projectname{TimeSRL}

\begin{document}

%


\title{TimeSRL: Generalizable Time-Series Behavioral Modeling via Semantic RL-Tuned LLMs --- A Case Study in Mental Health}

%

\author{Yuang Fan}
\email{yf2676@columbia.edu}
\orcid{}
\affiliation{%
  \institution{Columbia University}
  \country{USA}
}
\author{Lilin Xu}
\email{lx2331@columbia.edu}
\orcid{}
\affiliation{%
  \institution{Columbia University}
  \country{USA}
}
\author{Millie Wu}
\email{mw3209@columbia.edu}
\orcid{}
\affiliation{%
  \institution{Columbia University}
  \country{USA}
}
\author{Jingping Nie}
\email{jingping@unc.edu}
\orcid{}
\affiliation{%
  \institution{University of North Carolina at Chapel Hill}
  \country{USA}
}
\author{Qingyu Chen}
\email{qingyu.chen@yale.edu}
\orcid{}
\affiliation{%
  \institution{Yale University}
  \country{USA}
}
\author{Yuzhe Yang}
\email{yuzhey@ucla.edu}
\orcid{}
\affiliation{%
  \institution{University of California, Los Angeles}
  \country{USA}
}
\author{Zhuo Zhang}
\email{zz@cs.columbia.edu}
\orcid{}
\affiliation{%
  \institution{Columbia University}
  \country{USA}
}
\author{Xin Liu}
\email{xliucs@google.com}
\orcid{}
\affiliation{%
  \institution{Google}
  \country{USA}
}
\author{Subigya Nepal}
\email{sknepal@virginia.edu}
\orcid{}
\affiliation{%
  \institution{University of Virginia}
  \country{USA}
}
\author{Xiaofan Jiang}
\email{jiang@ee.columbia.edu}
\orcid{}
\affiliation{%
  \institution{Columbia University}
  \country{USA}
}
\author{Xuhai ``Orson'' Xu}
\email{xx2489@columbia.edu}
\orcid{}
\affiliation{%
  \institution{Columbia University}
  \country{USA}
}

%
\renewcommand{\shortauthors}{Fan et al.}
\renewcommand{\shorttitle}{TimeSRL: Generalizable Time-Series Behavioral Modeling via Semantic RL-Tuned LLMs}

%
\begin{abstract}
\input{0-abstract}
\end{abstract}

%
%
%

\begin{teaserfigure}
\centering
\includegraphics[width=0.9\textwidth]{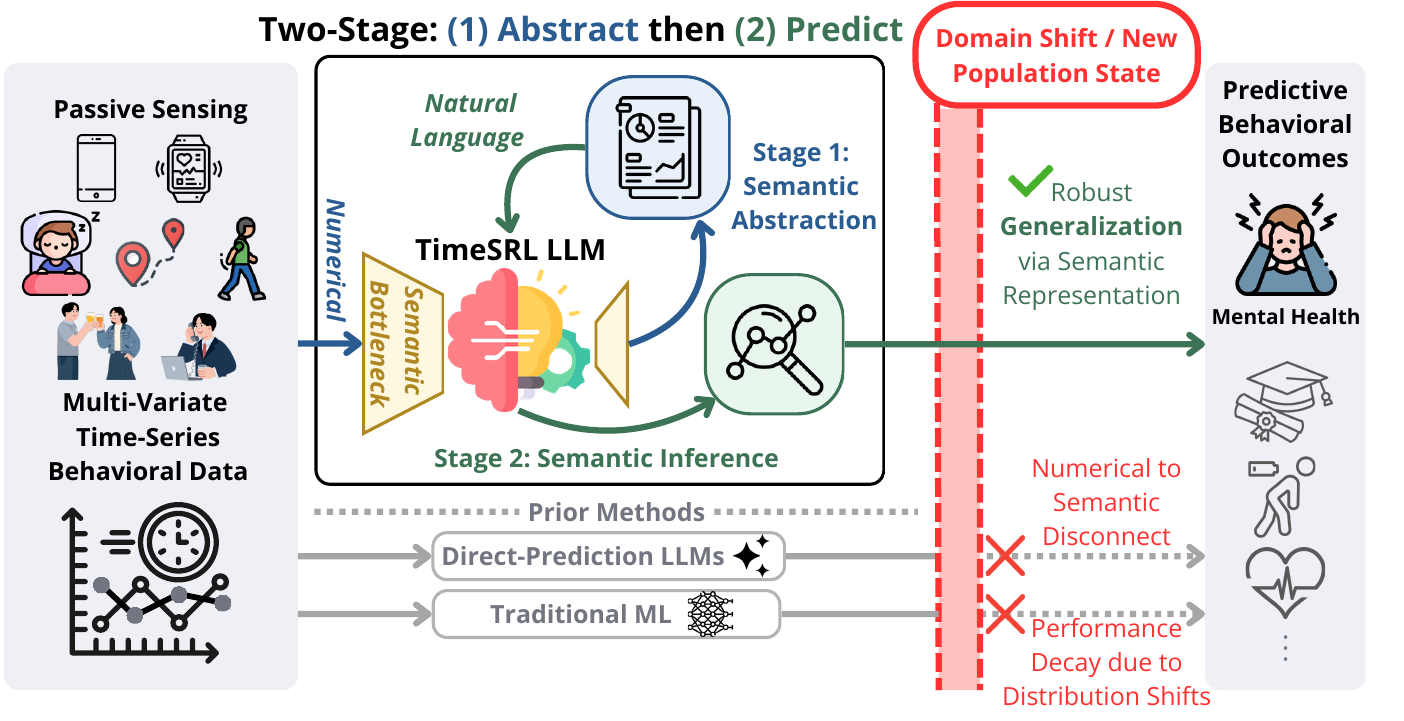}
\caption{\textbf{Overview of \projectname{}, a two-stage LLM framework for robust longitudinal behavioral time-series modeling, instantiated on behavioral health prediction.} While traditional ML models overfit numerical regularities and direct-prediction LLMs struggle with long numeric trajectories, \projectname{} addresses these distribution shift challenges by routing inference through an explicit \emph{semantic bottleneck}. In Stage 1, it abstracts raw numerical signals into natural-language behavioral descriptions; in Stage 2, it infers outcomes from this abstraction alone, enabling robust generalization across new populations. This paper focus on mental health prediction as a case study.
}
\label{fig:teaser}
\end{teaserfigure}

%
\maketitle

\input{1-introduction}

\input{2-background}

\input{3-methods}

\input{4-experiments}

\input{5-results}

\input{6-discussion}

\input{7-conclusion}


\section*{Acknowledgement The Use of AI}
We acknowledge that Generative AI tools (OpenAI ChatGPT and Anthropic Claude) were used solely to improve the quality of writing in this manuscript, including style, phrasing, and grammar polishing of author-written text. Typical prompts included variations of "improve the clarity and grammar of the following paragraph without changing its technical meaning" and "suggest more concise phrasing for this sentence." All AI-suggested edits were reviewed, verified, and approved by the authors, who take full responsibility for the content of the manuscript.                                                          
No part of the research contributions, including the problem formulation, methodology, algorithms, experimental design, results, analysis, and figures, was generated by Generative AI. The only exceptions are the core technical components of our method and experiments that explicitly involve generative AI and large language models as objects of study, where their use is described in detail in the Method and Experiments sections.            

\bibliographystyle{ACM-Ref-Format}
\bibliography{
bib/ref
}

\input{8-appendix}

\end{document}

%% file: variables.tex
\usepackage{balance}       
\usepackage{graphics}      
\usepackage{hyperref}
\usepackage{color}
\usepackage{booktabs}
\usepackage{textcomp}
\usepackage{enumerate}
\usepackage{xcolor}
\usepackage{lipsum}
\usepackage{makecell}
\usepackage{multicol}
\usepackage{multirow}
\usepackage{array}
\usepackage{verbatimbox}
\usepackage{enumitem}
\usepackage{amsmath}
\usepackage{stfloats}
\usepackage{amsthm}
\usepackage{listings}
\usepackage{caption} 
\usepackage{xspace}
\usepackage{algorithm}
\usepackage{algpseudocode}
\usepackage{tabularx}
\usepackage{arydshln}
\usepackage[bottom]{footmisc}
\usepackage{stackengine}
\usepackage{placeins}
\usepackage{graphicx}
\usepackage{subcaption} 
\usepackage{booktabs}
\usepackage{multirow}
\usepackage{amsmath}
\usepackage[inkscapelatex=false]{svg}

\usepackage[most]{tcolorbox}
\usepackage{tabularx}
\usepackage{booktabs}
\usepackage{xcolor}
\usepackage{array}
\definecolor{timesrlblue}{RGB}{43,92,145}
\definecolor{timesrlbluebg}{RGB}{238,244,250}
\definecolor{baselinegray}{RGB}{90,90,90}
\definecolor{baselinebg}{RGB}{245,245,245}
\definecolor{warnred}{RGB}{170,70,70}
\definecolor{warnredbg}{RGB}{252,240,240}
\definecolor{goodgreen}{RGB}{70,120,90}
\definecolor{goodgreenbg}{RGB}{240,247,242}

\usepackage[table]{xcolor}
\definecolor{LightGreen}{RGB}{198,239,206}
\definecolor{LightBlue}{RGB}{221,235,247}

\newcommand*{\eg}{\textit{e.g.},\xspace}
\newcommand*{\ie}{\textit{i.e.},\xspace}

\newcolumntype{L}[1]{>{\raggedright\let\newline\\\arraybackslash\hspace{0pt}}m{#1}}
\newcolumntype{C}[1]{>{\centering\let\newline\\\arraybackslash\hspace{0pt}}m{#1}}
\newcolumntype{R}[1]{>{\raggedleft\let\newline\\\arraybackslash\hspace{0pt}}m{#1}}

\makeatletter
\def\thickhline{%
  \noalign{\ifnum0=`}\fi\hrule \@height \thickarrayrulewidth \futurelet
   \reserved@a\@xthickhline}
\def\@xthickhline{\ifx\reserved@a\thickhline
               \vskip\doublerulesep
               \vskip-\thickarrayrulewidth
             \fi
      \ifnum0=`{\fi}}
\makeatother

\makeatletter
\def\thickhlinespace{%
  \addlinespace[1ex]
  \noalign{\ifnum0=`}\fi\hrule \@height \thickarrayrulewidth \futurelet
   \reserved@a\@xthickhline
   \addlinespace[1ex]
   }
\def\@xthickhlinespace{\ifx\reserved@a\thickhline
               \vskip\doublerulesep
               \vskip-\thickarrayrulewidth
             \fi
      \ifnum0=`{\fi}}
\makeatother

\newlength{\thickarrayrulewidth}
\newlength\Origarrayrulewidth

\algnewcommand{\IfThenElse}[3]{
  \State \algorithmicif\ #1\ \algorithmicthen\ #2\ \algorithmicelse\ #3}

\newenvironment{s_enumerate}{
\begin{enumerate}
  \setlength{\itemsep}{0pt}
  \setlength{\parskip}{0pt}
  \setlength{\parsep}{0pt}
}{\end{enumerate}}

\definecolor{downredcolor}{HTML}{e31a1c}
\definecolor{upgreencolor}{HTML}{33a02c}

\definecolor{DarkGreen}{HTML}{5DAC81}

%% file: 0-abstract.tex
Longitudinal passive sensing enables continuous health prediction, yet models often fail under cross-dataset distribution shifts. Traditional ML overfits cohort-specific artifacts, while Large Language Models (LLMs) struggle to reason reliably over long, heterogeneous time-series. 
We introduce \projectname{}, a two-stage LLM framework that routes predictions through an explicit \emph{semantic bottleneck}. The model first abstracts raw signals into high-level natural language, then predicts behavioral outcomes from these abstractions alone. This forces the model to reason over semantic concepts that we argue generalize better than raw numbers. We optimize this process end-to-end using Group Relative Policy Optimization (GRPO) with Reinforcement Learning from Verifiable Rewards (RLVR), learning outcome-aligned abstractions without gold intermediate annotations.
Instantiated on mental-health prediction, \projectname{} achieves state-of-the-art performance on a benchmark designed to stress-test cross-cohort generalization under a rigorous leave-one-dataset-out (LOSO) protocol, reducing mean absolute error (MAE) over strong non-LLM ML and LLM baselines by 3.1--10.1\% and 9.5--44.1\% for anxiety, and 3.2--9.6\% and 27.4--57.6\% for depression (all $p$s<0.05). \projectname{} significantly outperforms prior methods in cross-benchmark transfer across different sensing pipelines, rivaling its own within-domain performance without target-domain fine-tuning. These results demonstrate that semantic abstractions are reusable and point to a new direction for generalizable behavior modeling via RL-tuned LLMs.

%% file: 1-introduction.tex
\section{Introduction}
\label{sec:introduction}
Personal devices such as smartphones and wearables passively and continuously capture rich traces of everyday behavior, including mobility, sleep, device usage, physical activity, and routine regularity. For the ubiquitous computing community, these behavioral signals create the possibility of moving from sparse, retrospective assessment toward continuous, ecologically grounded understanding of human state in daily life. If modeled reliably, such passive behavioral sensing data could support a wide range of health and well-being applications, from longitudinal monitoring to earlier identification of meaningful behavioral change~\cite{wang_tracking_2018, rykov_digital_2021, moshe_predicting_2021, sun_challenges_2023, nepal_capturing_2024, yang2022artificial}. Numerous works have explored the potential of building artificial intelligence (AI) models using such longitudinal time-series data in various domains, such as mental health~\cite{saeb_mobile_2015, wang_tracking_2018, borelli_detection_2025, chikersal_detecting_2021, nie_ai_2022}, academic performance~\cite{wang_studentlife_2014, wang_smartgpa_2015}, sleep~\cite{abdullah_towards_2014, shuai2026osf, xu2026sleeplm}, and stress~\cite{sano_stress_2013, bogomolov_daily_2014, hovsepian_cstress_2015, mishra_continuous_2020}.

Yet, turning passive behavioral sensing into robust inference remains difficult. A central challenge is \textbf{generalization}. Human behavior varies substantially across individuals, populations, contexts, sensing pipelines, and historical periods~\cite{li2026hearts}. Multiple recent studies have demonstrated that models that perform well within one study often degrade when deployed on another cohort or under shifted behavioral distributions~\cite{xu_globem_2023, adler_machine_2022, meegahapola_generalization_2023}. In practice, this means that success on a single benchmark or population does not necessarily translate into reliable performance in the wild. For ubiquitous computing, where deployment conditions are inherently heterogeneous and evolving, robustness under distribution shift is therefore of critical importance.

This generalizability challenge is especially pronounced in longitudinal behavior modeling. Daily behavioral logs are noisy, high-dimensional, partially structured, and temporally entangled~\cite{xu2023globemdatasetmultiyeardatasets, nepal_capturing_2024, meegahapola_generalization_2023, adler_machine_2022}. The predictive signal often lies not in isolated values, but in higher-level patterns such as routine stability, disruption, withdrawal, recovery, volatility, and sustained change over time. Traditional machine-learning (ML) pipelines typically operate on engineered statistics or aggregated features, which can achieve reasonable performance within a dataset but often remain sensitive to cohort-specific numerical regularities~\cite{xu_globem_2023, adler_machine_2022, meegahapola_generalization_2023}. More recently, large language models (LLMs) have emerged as a promising alternative because they can potentially reason over higher-level behavioral semantics expressed in natural language rather than relying only on fixed feature mappings~\cite{kim_health-llm_2024, imran_llasa_2025, zheng_promind-llm_2025, xia_convergence_2025, li2026hearts}.
However, simply applying an LLM to raw behavioral data does not resolve the problem. Longitudinal sensing trajectories are long, heterogeneous, and weakly structured, making them difficult to reason over directly. Even when prompted with formatted data, LLMs may latch onto brittle numerical cues or produce plausible but weakly grounded explanations, making it challenging for cross-dataset generalization~\cite{zhang_sensorlm_2025, gu_radar_2025}. This raises a core question for ubiquitous behavioral modeling: \emph{how should behavioral time series be abstracted so that a model can reason over them more robustly with generalizability under real-world variability?}

We argue that robust inference should pass through an explicit \textbf{\emph{semantic bottleneck}}. Instead of predicting directly from raw behavioral data, the model should first extract semantic insights by transforming the trajectory into a compact natural-language abstraction that captures temporally extended and context-dependent patterns relevant for downstream inference. These abstractions should go beyond predefined features (e.g., regularity or variance) to encode unrestricted structure, such as evolving cross-signal dependencies, relative importance among behavioral signals, and individualized behavioral trajectory changes. Such patterns are inherently open-ended and task-adaptive, making them difficult to enumerate or capture through manual feature engineering. The prediction should then be made from this semantic abstraction, which we speculate is more robust to distribution shifts across datasets. This bottleneck serves two purposes. First, it encourages the model to operate on semantically meaningful behavioral structure rather than overfitting to brittle, dataset-specific numeric detail. Second, it makes the intermediate representation consequential rather than merely explanatory: the summary is not a post hoc rationale, but the representation through which the final prediction must pass.

However, such semantic bottleneck face a critical challenge: there is usually no gold supervision for what a useful intermediate behavioral summary should look like. In many ubiquitous sensing problems, we may have validated end-task labels, but not human-annotated semantic abstractions of the underlying trajectories. To address this challenge, we found that this setting is well-suited to reinforcement learning with verifiable outcomes (RLVR). Prior RLVR-style work suggests that ranking candidate trajectories by task reward can improve structured reasoning, not just final answers~\cite{shao_deepseekmath_2024, deepseek-ai_deepseek-r1_2025}. While such approaches have been studied primarily in more direct one-step generation settings, we apply the reward to the full two-step trajectory with the \emph{semantic bottleneck}. This encourages the model to produce summaries that preserve useful behavioral evidence and support accurate downstream prediction (\ie passing through the \emph{semantic bottleneck}), while discouraging summaries that are merely plausible but not decision-useful. In this way, reinforcement learning optimizes not only the final prediction, but also the semantic abstraction that supports it.

In this paper, we present \textbf{\projectname{}} (Time-series Semantic Reinforcement Learning), a two-stage LLM framework for robust and generalizable longitudinal behavioral time-series prediction. In the first stage, the model converts raw behavioral data into an unrestricted natural-language abstraction that includes but not limited to stability, anomalies, and temporal dynamics. In the second stage, it predicts the target outcome using only this abstraction, without access to the original numerical inputs. We then optimize this abstract-then-prediction process using Group Relative Policy Optimization (GRPO) as the RLVR method, so that the learned bottleneck is shaped by downstream utility rather than generic summarization quality. The result is an outcome-aligned semantic representation designed to support more robust inference under behavioral distribution shift.

We instantiate this framework in behavioral health, one of the most widely studied applications of passive sensing and the setting where cross-cohort generalization failures are most thoroughly documented~\cite{xu_globem_2023, adler_machine_2022, meegahapola_generalization_2023}, making mental-health prediction a natural testbed that pairs validated end-task labels (e.g., PHQ-4~\cite{kroenke_ultra-brief_2009}) suitable for RLVR with a benchmark explicitly designed to stress-test the robustness claim at the core of our approach. Using mental-health prediction as a case study, we show that \projectname{} achieves state-of-the-art generalization performance under a rigorous leave-one-dataset-out (LOSO) evaluation protocol. On anxiety prediction, \projectname{} reduces mean absolute error (MAE) by 3.1--10.1\% ($p$s<0.05) compared with strong non-LLM ML generalization baselines and by 9.5--44.1\% ($p$s<0.001) compared with LLM baselines. On depression prediction, the corresponding improvements are 3.2--9.6\% ($p$s<0.001) and 27.4--57.6\% ($p$s<0.001), respectively. 
We further demonstrate the robustness of \projectname{} by showing the improvement across multiple datasets and multiple backbone LLMs. Beyond within-benchmark generalization, \projectname{} also transfers across benchmarks with different sensing pipelines: trained on one benchmark and evaluated on another, it outperforms baselines on all tasks and splits and rivals its own within-benchmark performance without any target-domain fine-tuning, indicating that the learned semantic abstractions are reusable rather than benchmark-specific.

More broadly, our findings suggest a general modeling principle for ubiquitous computing: when raw longitudinal sensor data are difficult to generalize across people and contexts, explicitly learned semantic abstractions can provide a more stable interface between behavioral sensing and downstream inference. While we study behavioral health in this paper, the same perspective may be relevant to other ubiquitous computing problems that require robust reasoning over complex human behavioral trajectories.

Our contributions are threefold:
\begin{itemize}
    \item We introduce a two-stage semantic bottleneck framework for longitudinal behavioral time-series modeling that explicitly separates behavioral abstraction from downstream inference, and show how reinforcement learning can jointly optimize this bottleneck without gold intermediate annotations, using only validated end-task outcomes as supervision.
    \item We demonstrate that optimizing LLMs over outcome-aligned semantic abstractions yields more generalizable predictions under distribution shift, achieving state-of-the-art performance on a benchmark with pronounced cross-cohort distribution shift and outperforming baselines by 3.1--57.6\% with statistical significance ($p$s$<$0.05).
    \item We further validate our framework's robustness across multiple datasets and LLM backbones, and show that its learned semantic abstractions transfer across benchmarks with different sensing pipelines without target-domain fine-tuning, pointing to a broader framework for generalizable behavior modeling in ubiquitous computing.
\end{itemize}

%% file: 2-background.tex
\section{Related Work}
\label{sec:related_work}



Prior work relevant to \projectname{} spans three connected threads.
First, passive sensing from mobile and wearable devices has become an important basis for modeling everyday behavior across ubiquitous computing applications, with mental health serving as one important instantiation.
This literature has shown both the promise of longitudinal behavioral modeling and the persistent difficulty of generalizing across cohorts, contexts, and sensing conditions (Sec.~\ref{sub:related:mobile_sensing}).
Second, LLMs have recently been explored as a flexible interface for passive sensing data, raising the possibility of reasoning over higher-level behavioral semantics rather than only fixed numerical features (Sec.~\ref{sub:related:llm_sensing}).
Third, reinforcement learning has emerged as a useful tool for tuning LLMs toward task-specific objectives, including settings where useful intermediate reasoning must be learned indirectly from end-task outcomes (Sec.~\ref{sub:related:rl_llm}).

\subsection{Mobile Sensing for Behavior Modeling, Behavioral Health, and the Generalizability Challenge}
\label{sub:related:mobile_sensing}
\textbf{Behavior modeling from passive sensing.}
Smartphones and wearables continuously record physical, social, and routine signals---mobility, sleep, communication, app usage, location variance, and circadian rhythm---that together provide a digital phenotype of everyday behavior~\cite{nie_spiders+_2021, mohr_personal_2017, harari_using_2016}, spanning early phone-based inferences of behavioral state~\cite{likamwa_moodscope_2013, rabbi_mybehavior_2015} and multi-year longitudinal cohort deployments~\cite{wang_studentlife_2014, nepal_capturing_2024}.
Because these signals are passive, longitudinal, and ecologically grounded, they support behavior modeling at a temporal granularity that questionnaires cannot easily reach.
This broader sensing paradigm has been used across domains including sleep~\cite{abdullah_towards_2014, min_toss_2014}, stress~\cite{sano_stress_2013, bogomolov_daily_2014, hovsepian_cstress_2015, sano_identifying_2018}, academic performance~\cite{wang_smartgpa_2015, wang_studentlife_2014}, health-risk behaviors such as alcohol use~\cite{bae_detecting_2017, fan_exploring_2024}, and social functioning~\cite{doryab_identifying_2019}.
A central and widely studied instantiation of this paradigm is mental and behavioral health, where passively sensed behavioral features have been associated with and used to predict depression~\cite{saeb_mobile_2015, wang_tracking_2018, rykov_digital_2021, sun_challenges_2023, nepal_moodcapture_2024}, anxiety and mood~\cite{moshe_predicting_2021, boukhechba_predicting_2018, servia-rodriguez_mobile_2017}, and symptom trajectories in serious mental illnesses such as schizophrenia~\cite{wang_crosscheck_2016, wang_predicting_2017, adler_predicting_2020}, with systematic reviews synthesizing these efforts~\cite{amin_use_2025}.
Early work in this area often formulated the task as supervised learning over engineered or aggregated features, using linear regression, support vector machines, gradient-boosted trees, multilayer perceptrons, or temporal neural models~\cite{wang_studentlife_2014, canzian_trajectories_2015, saeb_mobile_2015, jaques_predicting_2015}, with later refinements introducing contextually filtered and routine-aware features~\cite{chikersal_detecting_2021, xu_leveraging_2019}.

\noindent\textbf{The cross-cohort generalization challenge.}
A recurring finding across this literature is that models that fit one cohort well may degrade substantially on another.
Behavioral signals are sensitive to population, geographic and institutional context, sensing-pipeline differences, individual heterogeneity, and large-scale temporal shifts such as the COVID-19 pandemic, all of which can alter the joint distribution over numerical features~\cite{adler_machine_2022, meegahapola_generalization_2023, xu_globem_2023, sun_challenges_2023, mack_mental_2021}.
Cross-cohort evaluations on several benchmarks make this concrete: methods that achieve competitive within-study performance can approach near-baseline performance when transferred to a new dataset~\cite{xu_globem_2023, adler_machine_2022, meegahapola_generalization_2023, pillai_investigating_2024}, and recent reviews identify external validation as a major open problem in the field~\cite{amin_use_2025}.
Recent state-of-the-art systems have also begun to respond to this challenge in different ways.
ReOrder~\cite{xu_globem_2023} targets cross-dataset robustness directly through temporally structured representation learning and therefore provides an especially relevant generalization-oriented baseline.
Borelli et al.~\cite{borelli_detection_2025} and Ahmed et al.~\cite{ahmed_explainable_2025} are not primarily framed around cross-cohort transfer, but their stronger temporal aggregation, multimodal feature construction, and domain-informed signal extraction may still help reduce brittleness relative to simpler feature pipelines.
In our experiments, these systems serve as strong non-LLM baselines.

However, these approaches still pursue robustness by changing the model class or feature pipeline while treating a fixed numerical feature vector as the predictive representation.
This leaves open a different possibility that this work aims to explore: shifting the predictive representation from fixed numerical features to a higher-level abstraction of behavioral trajectories, potentially mediated through language and used as the basis for downstream prediction.

\subsection{Large Language Models for Sensor Data}
\label{sub:related:llm_sensing}
\textbf{LLMs as a general interface to sensor data.}
A growing body of work uses LLMs to interpret, describe, or reason over sensor streams.
One direction develops sensor--language foundation models that align multivariate sensor signals to text via large-scale pretraining, including SensorLM~\cite{zhang_sensorlm_2025}, SensorLLM~\cite{li_sensorllm_2025}, SleepLM~\cite{xu2026sleeplm}, and the Large Sensor Model that scales wearable foundation models to tens of millions of hours of multimodal data~\cite{narayanswamy_scaling_2024}.
A complementary direction uses LLMs as semantic translators~\cite{nepal_mindscape_2024}.
DailyLLM~\cite{tian_dailyllm_2025} and AWARE Narrator~\cite{zhang_aware_2024} generate natural-language behavioral diaries from smartphone logs.
Time-LLM reprograms language models for general time-series forecasting~\cite{jin_time-llm_2024}, and PHIA~\cite{merrill_transforming_2026} casts wearable data analysis as an LLM-agent task with code execution and retrieval.
Taken together, these efforts suggest that sensor data can be mapped into language-like representations that LLMs can manipulate productively.

\noindent\textbf{LLMs for behavioral-health prediction.}
A second thread applies LLMs more directly to behavioral-health and well-being prediction~\cite{englhardt_classification_2024, xu_mental-llm_2024, nie_llm-based_2025, heydari_anatomy_2025, li2026hearts}.
Health-LLM~\cite{kim_health-llm_2024} prompts LLMs over wearable-derived features for multi-task health prediction.
LLaSA~\cite{imran_llasa_2025} fine-tunes a multimodal LLM for human activity analysis from smartphone and wearable signals.
ProMind-LLM~\cite{zheng_promind-llm_2025} combines domain-specific pretraining with chain-of-thought reasoning over sensor data for proactive mental-health care.
These studies suggest that LLMs may help capture higher-level semantic patterns that are difficult to specify through fixed feature mappings alone.
At the same time, they also expose a recurring difficulty: pretrained LLMs are not naturally optimized for long, heterogeneous, weakly structured numeric trajectories, and can still rely on brittle numerical cues or produce plausible but weakly grounded explanations~\cite{gu_radar_2025, xia_convergence_2025}.

Across both threads, language is typically treated as an output medium or retrieval interface. In contrast, our approach proposes using language as an \emph{intermediate} representation for downstream prediction. By routing predictions through an inspectable intermediate state, this design shares a core intuition with \emph{concept bottleneck models} (CBMs)~\cite{koh_concept_2020} in the broader machine learning literature. However, while traditional CBMs rely on a fixed, human-specified concept vocabulary trained with explicit intermediate supervision, \projectname{} instantiates this bottleneck as an open-ended, natural-language abstraction. Because predictive behavioral structures are highly difficult to enumerate in advance, we learn this semantic abstraction entirely via RL from end-task labels, without gold intermediate supervision.

\subsection{Reinforcement Learning for LLM Tuning}
\label{sub:related:rl_llm}
\textbf{From RLHF to RLVR.}
Reinforcement learning has become a central paradigm for shaping LLM behavior beyond what supervised fine-tuning alone can typically achieve.
Early work used RL from human feedback (RLHF) to align models with human preferences via reward models trained on pairwise rankings~\cite{ouyang_training_2022}.
More recently, Reinforcement Learning with Verifiable Rewards (RLVR) replaces the learned reward model with a deterministic verifier---a rule, unit test, or correctness check---and has driven strong gains on reasoning benchmarks in models such as T\"ulu~3, DeepSeek-Math, DeepSeek-R1, and Qwen3~\cite{lambert_tulu_2025, shao_deepseekmath_2024, deepseek-ai_deepseek-r1_2025, qwen3technicalreport}.
Group Relative Policy Optimization (GRPO), introduced by DeepSeek-Math~\cite{shao_deepseekmath_2024}, is a critic-free variant that ranks groups of sampled trajectories against their average reward and has become a common backbone for RLVR-style training.

\noindent\textbf{RL can shape intermediate reasoning, not only final answers.}
A key observation from this line of work is that outcome-based rewards do not only improve the final answer; they can also reshape the intermediate reasoning trajectories that models produce, encouraging steps that are more coherent, decision-useful, and aligned with the final objective~\cite{deepseek-ai_deepseek-r1_2025, wen_reinforcement_2025}.
This observation is relevant to our setting because in behavioral sensing, validated end-task labels are often available, whereas human-annotated intermediate behavioral summaries are generally not.
Any attempt to learn a useful abstraction therefore needs to rely largely on indirect supervision through the final outcome.
At the same time, RLVR-style optimization has been studied primarily in domains with relatively short and more readily verifiable answers---such as math, code, and structured question answering---rather than in long, noisy, multivariate behavioral sensing tasks where the intermediate representation is itself a free-form natural-language summary of a trajectory.
This makes RL-based tuning a plausible but still underexplored opportunity for our problem setting.
Our work explores this opportunity in the behavioral sensing setting.

%% file: 3-methods.tex
\section{Methods}
\label{sec:methods}
This section formalizes the prediction problem (Sec.~\ref{sec:method:problem}), introduces the \projectname{} semantic-bottleneck framework (Sec.~\ref{sec:method:overview}), details its two stages (Sec.~\ref{sub:stage1}, Sec.~\ref{sub:stage2}), and describes the joint GRPO optimization that ties them (Sec.~\ref{sub:rl}).

\subsection{Problem Formulation}
\label{sec:method:problem}
To predict longitudinal behavior, each sample consists of a multi-day window of daily behavioral sensing data, including signals such as activity, sleep, device usage, communication, and mobility, paired with an application-specific target label. Let $X=\{x_1,\dots,x_T\}$ denote a sequence of multivariate behavioral data over $T$ days, where $x_t \in \mathbb{R}^d$ is the aggregated sensing feature vector for day $t$ with a dimension of $d$, and let $y$ denote the downstream target (\eg a PHQ-4 anxiety score). The conventional formulation learns a direct predictor $f(X) \rightarrow y$.

However, as evident by prior work~\cite{xu_globem_2023, adler_machine_2022, meegahapola_generalization_2023, tseng_using_2020}, direct prediction from raw numerical features is difficult to generalize across cohorts, sensing environments, and time periods. Behavioral data vary substantially in scale and distribution, causing models to latch onto dataset-specific numerical regularities that break under distribution shift. The predictive signal in longitudinal sensing often lies not in isolated feature values, but in higher-order structure that spans multiple signals and unfolds over time---evolving cross-signal dependencies, shifts in the relative dominance among behavioral channels, and individualized trajectory changes defined against a person's own prior routine. Such structure is inherently open-ended and task-adaptive: the informative pattern is rarely a pre-specifiable feature but a composition whose relevance emerges only in context, making it difficult to enumerate or capture through manual feature engineering. This motivates an approach that explicitly separates the extraction of such structure from the downstream prediction itself.

\begin{figure}
    \centering
    \includegraphics[width=0.9\textwidth]{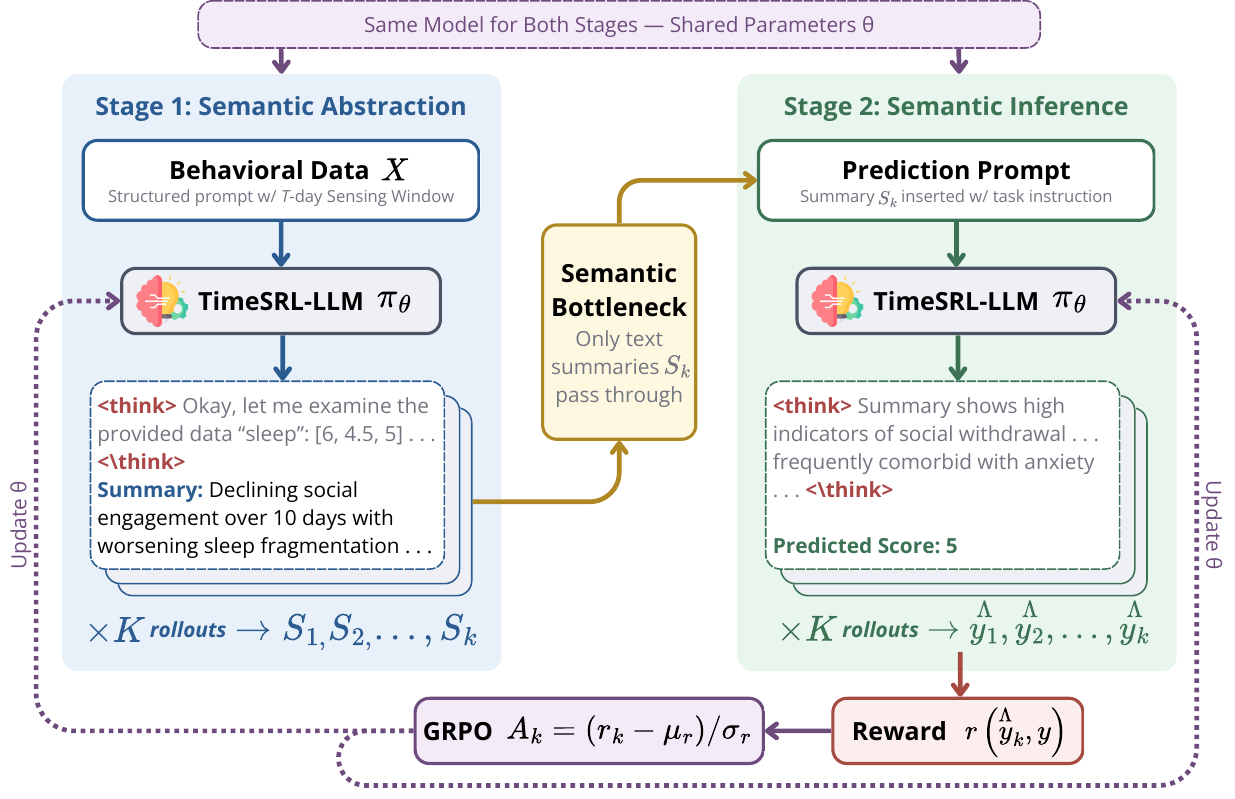}
    \caption{\textbf{The Two-Stage GRPO Tuning Pipeline for \projectname{}.} The proposed architecture uses the same model for both stages. In Stage~1, the \projectname{}-LLM is given a prompt with behavioral data to examine the numerical data and summarize the findings. The model leverages an explicit reasoning process to generate \#K semantic abstracted summaries. Next, only the generated summaries are extracted, passing through the semantic bottleneck, and inserted into a prediction prompt. In Stage~2, the \projectname{}-LLM executes predictive outcome inference based on the summaries, producing \#K outcome predictions. Finally, \#K scores are extracted and evaluated using reward function(s) to compute a GRPO update that optimizes the base \projectname{}-LLM.}
    \label{fig:grpo_pipeline}
    \vspace{-\baselineskip}
\end{figure}

\subsection{\projectname{} Overview}
\label{sec:method:overview}
\projectname{} addresses this challenge by introducing an explicit \emph{semantic bottleneck} between raw behavioral data and downstream inference. Instead of mapping numerical features directly to a prediction, the model first abstracts the behavioral trajectory into a compact natural-language summary, and then predicts the target from that summary alone:
\begin{equation}
S = g_{\theta}(X), \qquad \hat{y} = h_{\theta}(S),
\end{equation}
where $S$ is a natural-language abstraction of the behavioral data window $X$ and $\hat{y}$ is the predicted target. The summary $S$ is designed to capture higher-level behavioral structure, including temporally extended and inter-feature evolving patterns relevant for downstream inference that go beyond predefined features. Operating in natural language allows the bottleneck to leverage the reasoning and world knowledge that a pretrained LLM already possesses, so the model can recognize and describe behavioral constructs without needing to learn these concepts from sensor data alone. By forcing prediction to pass through this semantic bottleneck, the model operates on semantic patterns rather than brittle numerical magnitudes, making inference less sensitive to distribution shift.

As shown in Fig.~\ref{fig:grpo_pipeline}, both stages are executed by a single shared language model. In Stage~1 (Sec.~\ref{sub:stage1}), the model reads a structured representation of the behavioral window and produces a natural-language summary of temporal patterns and insights. In Stage~2 (Sec.~\ref{sub:stage2}), it predicts the target outcome from that summary alone. Because supervision is available only for the final prediction, not for the intermediate abstraction, we jointly optimize both stages with Group Relative Policy Optimization (Sec.~\ref{sub:rl}), so that the summarization policy is shaped directly by downstream predictive utility. Sharing parameters across stages is what makes this joint optimization possible: the reward signal propagates through the same weights that generate the summary, so Stage~1 is trained through Stage~2's predictive success without requiring the gold intermediate supervision that separate models would need.

\subsection{Stage 1: Semantic Abstraction from Behavioral Data}
\label{sub:stage1}
Raw behavioral data are high-dimensional, numerical, and only weakly structured, making them difficult for language models to reason over directly. To make behavioral trajectories accessible while preserving their temporal structure, Stage~1 transforms each input window into a structured prompt and asks the model to produce a behavioral summary.

Concretely, we preprocess each multi-day window into a structured prompt template that organizes daily observations, maps system-level feature names to descriptive labels, and converts raw units into interpretable forms. The prompt then asks the model to summarize the trajectory's behavioral structure in free-form natural language rather than return a fixed set of predefined statistics. The specific prompt design and preprocessing choices used in our experiments are described in Sec.~\ref{sec:experiments} and illustrated in Fig.~\ref{fig:prompt_turn1}.

When the Stage~1 model supports explicit reasoning traces, we require any intermediate reasoning to appear inside \texttt{<think>...</think>} or other functionally equivalent tags, and extract only the text following the closing tag as the Stage~1 summary. If the reasoning tags are missing, the summary is treated as empty and receives zero reward. This separation ensures that internal reasoning remains distinct from the semantic representation passed to Stage~2.

The goal of Stage~1 is not to restate the raw data, but to distill behavior into compositional patterns that are more likely to transfer across settings. The predictive signal in longitudinal sensing typically resides in the temporal interaction among multiple behavioral streams rather than in any individual channel. Consider two students with near-identical weekly sensor profiles---both show reduced daytime mobility, elevated late-night screen use, and increasingly fragmented sleep. The same surface evidence, however, may reflect fundamentally different latent states: one student drifting into depressive withdrawal, the other adapting to a compressed exam week before returning to baseline. Distinguishing them requires reasoning about the \emph{ordering} of changes across streams (does mobility decline precede or follow sleep disruption?), their \emph{persistence} (is late-night activity sporadic or sustained over weeks?), and their \emph{coherence} with contextual regularities such as prior routines and academic rhythms. No per-feature normalization recovers this distinction, and manually enumerating the relevant orderings, co-occurrences, and conditional dependencies across behavioral streams is intractable. By abstracting multi-signal dynamics into semantic descriptions, Stage~1 offloads this enumeration to the LLM's capacity for open-ended compositional reasoning, preserving relational and temporal structure while reducing sensitivity to cohort-specific scales and distributions.

\subsection{Stage 2: Predictive Inference from Semantic Abstraction}
\label{sub:stage2}
Stage~2 enforces the semantic bottleneck by predicting the target outcome solely from the natural-language summary $S$ produced by Stage~1, without access to the original behavioral data. This restriction ensures that the model can succeed only by reasoning over behavioral concepts expressed in language, rather than memorizing numerical shortcuts.

The Stage~2 prompt presents the summary and asks the model to produce a prediction in a task-specific format. During post-processing, we first discard any content before the closing \texttt{</think>} tag and then extract the final prediction using a fixed parser tailored to the target task. Outputs that violate the required format or cannot be parsed receive zero reward. This ensures that the optimization evaluates only well-formed predictions and prevents the model from being rewarded for formatting artifacts.

Because the same model parameters are shared across both stages, the supervision applied to Stage~2 also shapes the Stage~1 summarization policy. Summaries that omit critical context, overstate isolated outliers, or distort the trajectory are less likely to support correct prediction, and therefore receive weaker reinforcement during training. This creates a direct feedback path from prediction quality back to the semantic abstraction, which we formalize in the next section.

\input{figs/grpo_algo}

\subsection{Joint Optimization with GRPO}
\label{sub:rl}
The two-stage framework requires training without gold-standard intermediate annotations. We have validated end-task labels (\eg PHQ-4 scores for Stage 2) but no human-annotated targets specifying what a useful behavioral summary should contain (for Stage 1). This rules out supervised fine-tuning for Stage~1 alone and calls for an optimization method that can propagate task-level supervision through the full pipeline.

We address this with Group Relative Policy Optimization (GRPO)~\cite{shao_deepseekmath_2024, deepseek-ai_deepseek-r1_2025}, described in Algorithm~\ref{alg:timesrl}. For each training instance $(X,y)$, the model samples $K$ complete two-stage trajectories,
\begin{equation}
\tau_k = \bigl(S_k,\hat{y}_k\bigr), \qquad k=1,\dots,K,
\end{equation}
where $S_k$ is a candidate semantic abstraction from Stage~1 and $\hat{y}_k$ is the corresponding Stage~2 prediction. Each trajectory receives a scalar reward based on the parsed prediction, and GRPO updates the policy to prefer trajectories with higher relative reward within the sampled group. The optimization objective is
\begin{equation}
\max_{\theta}\; \mathbb{E}_{\tau\sim\pi_{\theta}(\cdot|X)}
\bigl[r(\tau)\bigr]
-\beta\,\mathrm{KL}\!\left(\pi_{\theta}\,\|\,\pi_{\text{ref}}\right),
\end{equation}
where $\pi_{\theta}$ is the current policy, $\pi_{\text{ref}}$ is the reference pretrained model, $r(\tau)$ is the task reward, and the Kullback–Leibler divergence (KL) term~\cite{joyce_kullback-leibler_2011} regularizes the update to preserve linguistic fluency and prevent degenerate drift.

The reward function $r(\tau)$ depends on the downstream task. In general, it favors trajectories whose final prediction is both correct and well-formed, and may combine a primary accuracy term with structure-aware components such as ordinal consistency or threshold-aware bonuses. The specific reward design in our experiments is described in Sec.~\ref{sec:experiments}.

Although reward is computed only from the final prediction, GRPO updates the probability of the entire two-stage trajectory. This creates a closed feedback loop: semantic abstractions that preserve task-relevant behavioral structure lead to more accurate predictions and receive stronger reinforcement, while abstractions that are incomplete or misleading produce worse predictions and are reinforced less. In effect, Stage~1 is learned through the predictive usefulness of the information it preserves for Stage~2.

During training, we sample $K$ trajectories per instance to enable relative ranking within GRPO. At inference time, the model generates a single summary and a single prediction for each behavioral window. The resulting prediction remains grounded in the intermediate abstraction, which also serves as an inspectable account of the behavioral evidence used for inference.

%% file: figs/grpo_algo.tex
\begin{algorithm}[t]
\caption{\projectname{} Training via Two-Stage Semantic Bottleneck and GRPO}
\label{alg:timesrl}
\small
\begin{algorithmic}[1]
\Require Training set $\mathcal{D}=\{(X_i,y_i)\}_{i=1}^N$, current policy $\pi_\theta$, reference policy $\pi_{\mathrm{ref}}$
\Require Rollout count $K$, task-specific reward function $r(\hat{y}, y)$
\For{each training iteration}
    \State Sample a minibatch $\mathcal{B} \subset \mathcal{D}$
    \For{each $(X,y) \in \mathcal{B}$}
        \For{$k=1, \dots, K$}
            \State \textit{\% Stage 1: Semantic Abstraction Generation}
            \State Sample completion $o_1^{(k)} \sim \pi_\theta(\cdot \mid X)$
            \State Extract semantic summary $S_k$ from $o_1^{(k)}$ (excluding \texttt{<think>} reasoning)

            \State \textit{\% Stage 2: Prediction from Semantic Abstraction}
            \State Sample completion $o_2^{(k)} \sim \pi_\theta(\cdot \mid S_k)$
            \State Parse prediction $\hat{y}_k$ from $o_2^{(k)}$

            \State \textit{\% Reward Computation}
            \If{format invalid \textbf{or} $\hat{y}_k$ unparseable}
                \State $r_k \gets 0$
            \Else
                \State $r_k \gets r(\hat{y}_k, y)$
            \EndIf
            \State Store trajectory $\tau_k=(X, o_1^{(k)}, S_k, o_2^{(k)})$ and reward $r_k$
        \EndFor

        \State \textit{\% Group Relative Policy Optimization (GRPO)}
        \State Compute group baseline $\mu_r = \frac{1}{K}\sum_{k=1}^K r_k$ and standard deviation $\sigma_r$
        \State Compute normalized advantages $A_k = (r_k - \mu_r) / (\sigma_r + \epsilon)$ for $k=1, \dots, K$
        \State Accumulate objective $\mathcal{L}_{\mathrm{GRPO}}$ using $A_k$ and KL divergence $\mathbb{D}_{\mathrm{KL}}(\pi_\theta \parallel \pi_{\mathrm{ref}})$
    \EndFor
    \State Update parameters $\theta$ via gradient ascent on $\mathcal{L}_{\mathrm{GRPO}}$
\EndFor
\end{algorithmic}
\end{algorithm}

%% file: 4-experiments.tex
\section{Experimental Setup}
\label{sec:experiments}
Below we describe the datasets, task instantiation, baseline methods, and evaluation protocol. Baseline comparisons and the design of each robustness and ablation experiment are introduced alongside their corresponding results in Sec.~\ref{sec:results}.

\begin{figure}
    \centering
    \includegraphics[width=1\textwidth]{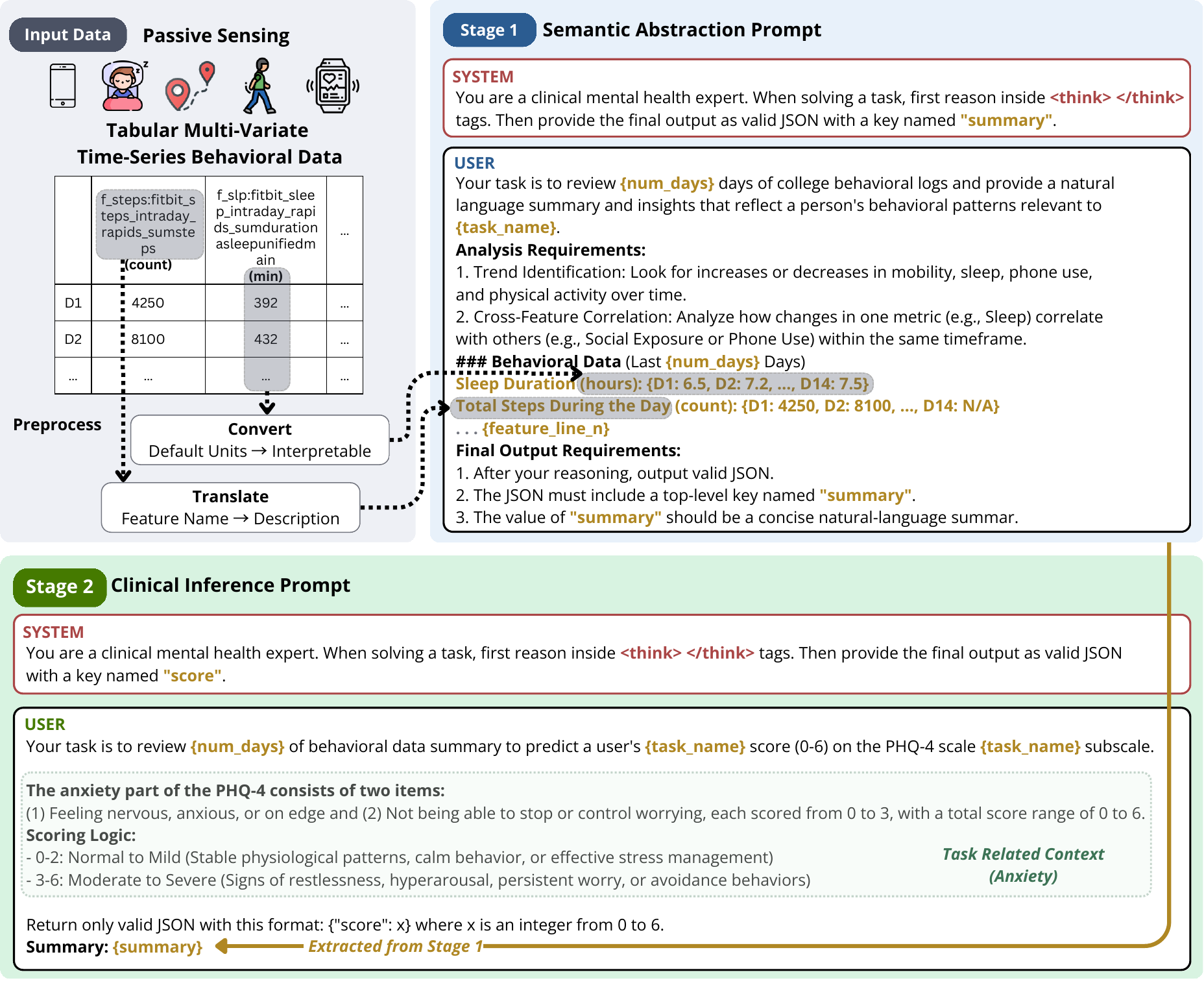}
    \caption{\textbf{Example of the Two Stage Prompting used in the task of mental-health prediction.} Starting from 14 days of tabular multi-variate time-series behavioral data, \projectname{} first constructs a semantic abstraction prompt in Stage~1 by organizing the data into a structured template, translating system feature names into descriptive labels, and converting raw sensor units into interpretable formats. This prompt guides the model to summarize behavioral patterns, trajectories, and notable fluctuations over the past two weeks in natural language. The resulting summary is then passed to the semantic inference prompt in Stage~2, which provides task-specific context and asks the model to infer the final mental-health score.}
    \Description{Workflow diagram showing steps to preprocess behavioral logs into a semantic abstraction prompt for LLMs.}
    \label{fig:prompt_turn1}
\end{figure}

\subsection{Datasets and Benchmarks}
\label{sub:experiment:dataset}
We instantiate the general formulation in Sec.~\ref{sec:method:problem} on behavioral mental health prediction. This domain offers a rigorous testbed for our framework: prediction targets are measured by well-established questionnaire used to support clinical diagnosis, inference requires reasoning over multi-day behavioral trajectories, and cross-cohort generalization remains a persistent open challenge~\cite{xu_globem_2023, adler_machine_2022, meegahapola_generalization_2023}.

We evaluate \projectname{} on two longitudinal behavioral sensing datasets that differ in distribution shift severity: GLOBEM~\cite{xu2023globemdatasetmultiyeardatasets}, a benchmark explicitly designed for cross-cohort generalization with substantial inter-subset heterogeneity, and College Experience~\cite{nepal_capturing_2024}, a more homogeneous dataset that provides a complementary evaluation under milder shift.

\textbf{GLOBEM~\cite{xu2023globemdatasetmultiyeardatasets}.}
GLOBEM is a longitudinal behavioral sensing benchmark containing passive behavioral data paired with weekly PHQ-4~\cite{kroenke_ultra-brief_2009} self-assessments. The dataset comprises four subsets (DS1--DS4) collected from different populations and time periods; we exclude DS1 because PHQ-4 assessments were not collected consistently. Each label is paired with behavioral features from the preceding 14 days, matching the questionnaire's retrospective window. We use a set of behavioral features commonly adopted in prior work~\cite{xu_leveraging_2021, xu_globem_2023, borelli_detection_2025, ahmed_explainable_2025} (see Appendix~\ref{app:feature_sets}), yielding 1{,}448 samples from DS2, 767 from DS3, and 1{,}264 from DS4. The subsets are subject-disjoint and calendar-time-disjoint, spanning noticeably different collection conditions including pre- and post-COVID periods.

\textbf{College Experience~\cite{nepal_capturing_2024}.}
College Experience contains passive behavioral data paired with weekly PHQ-4~\cite{kroenke_ultra-brief_2009} self-assessments from a single college cohort. To support cross-split evaluation, we partition the dataset into three disjoint subsets (named DS2, DS3, and DS4 for consistency), yielding 1{,}608 samples from DS2, 1{,}704 from DS3, and 1{,}800 from DS4. These splits are subject-disjoint and calendar-time-disjoint, but because they are drawn from the same overall cohort (i.e. all students enrolled in 2019), the distribution shift across splits is milder than in GLOBEM, providing a complementary evaluation condition.

\subsection{Prediction Tasks: PHQ-4 Anxiety and Depression}
\label{sub:experiment:task}
\input{figs/phq4_table}
The prediction target is derived from the PHQ-4~\cite{kroenke_ultra-brief_2009} questionnaire, a four-item self-assessment screening instrument whose structure is shown in Fig.~\ref{tab:phq4}. The PHQ-4 comprises two subscores: Anxiety (GAD-2\cite{spitzer_brief_2006}, items 1--2) and Depression (PHQ-2~\cite{kroenke_phq-9_2001}, items 3--4), each ranging from $0$ to $6$. For each sample, the target $y$ is one of these subscores, paired with behavioral data from the preceding 14 days to match the questionnaire's two-week retrospective window. We evaluate anxiety and depression as separate prediction tasks, as they capture distinct symptom dimensions that may relate to different behavioral patterns.

\subsection{Task-Specific Reward Design}
\label{sub:task_reward}
PHQ-4 subscores are ordinal integers in $\{0,\dots,6\}$, so an exact-match reward would be uninformative whenever the prediction is close but not perfectly correct. We therefore design a reward that provides dense gradient signal across the full score range.

Stage~2 is instructed to output a single integer in $\{0,\dots,6\}$ as the final prediction. During post-processing, we discard any content before \texttt{</think>} and extract the predicted score using a fixed regular expression over the required answer tags. Outputs that violate the format or cannot be parsed receive zero reward.

For valid outputs, let $p$ denote the predicted score and $y$ the ground-truth label. The reward is
\begin{equation}
r(p,y)=\exp\!\left(-\frac{(p-y)^2}{2\sigma^2}\right),
\end{equation}
which decays smoothly with squared error. This formulation encourages the model to move toward the correct score even from distant initial predictions, while the format constraint ensures adherence to the required output structure.

\subsection{Evaluation Protocol}
All experiments use held-out split generalization: models are trained on a subset of splits and evaluated on an entirely unseen split whose subjects and collection period differ from training, testing robustness under distribution shift rather than performance under random interpolation.

For GLOBEM, we adopt a leave-one-subset-out (LOSO) protocol over DS2--DS4: in each fold, two subsets are used for training and the remaining subset is held out for evaluation. For College Experience, we apply an analogous LOSO protocol over the three constructed splits. 

Performance is measured by mean absolute error (MAE) between predicted and ground-truth PHQ-4 subscores, averaged across evaluation folds. Statistical significance is assessed via paired bootstrap testing with 5{,}000 resamples on the pooled test predictions.

\subsection{Baselines}
\label{sub:baselines}
To disentangle the effects of model class, model scale, and the proposed training framework, we compare \projectname{} against three categories of baselines.

\textbf{Traditional feature-based models.}
Linear Regression, Support Vector Regression (SVR), XGBoost, and a multilayer perceptron (MLP), all trained on fixed behavioral features, represent standard non-LLM approaches for passive sensing mental-health prediction.

\textbf{State-of-the-art non-LLM behavioral modeling methods.}
We compare against strong recent methods for passive sensing behavioral prediction. Borelli et al.~\cite{borelli_detection_2025} fuse multiple device streams through systematic multimodal temporal aggregation, and Ahmed et al.~\cite{ahmed_explainable_2025} introduce a domain-informed feature extraction pipeline that captures fine-grained behavioral patterns. ReOrder~\cite{xu_globem_2023} targets cross-dataset robustness through temporally structured representation learning and serves as a strong generalization-oriented benchmark.

\textbf{LLM-based baselines.}
We anchor the LLM comparisons at two ends of the model-scale spectrum: GPT-5.0~\cite{openai_gpt5_2026}, a frontier model whose strong general reasoning provides an upper reference for what pre-trained knowledge alone can achieve, and Qwen3-4B~\cite{qwen3technicalreport}, a compact open-weight model that serves as the primary backbone for \projectname{} training.
Both backbones support native chain-of-thought reasoning, which is enabled in all configurations. The direct-prompting configurations instantiate the single-stage LLM inference paradigm used in prior work such as Health-LLM~\cite{kim_health-llm_2024} and ProMindLLM~\cite{zheng_promind-llm_2025}.

\subsection{Implementation Details}
\projectname{} is implemented with \texttt{ms-swift}~\cite{zhao_swiftscalable_2025} and the Hugging Face \texttt{trl} library for GRPO-based reinforcement learning. We apply LoRA (rank 16, $\alpha{=}32$) for parameter-efficient fine-tuning of Qwen3-4B. GRPO training uses $\sigma=1.2$, rollout size $K{=}8$, and input batch size 32, yielding an effective batch of 256 trajectories per update, with a cosine-annealed learning rate from $5{\times}10^{-5}$. Training runs for three epochs on four NVIDIA H100 GPUs; we select the checkpoint at which reward standard deviation stabilizes.

%% file: figs/phq4_table.tex
\begin{table}[t]
\centering
\small
\renewcommand{\arraystretch}{1.3}
\caption{\textbf{Structure of the PHQ-4~\cite{kroenke_ultra-brief_2009} questionnaire.} The four items are organized into two subscores: Anxiety (GAD-2~\cite{spitzer_brief_2006}, items 1--2) and Depression (PHQ-2~\cite{kroenke_phq-9_2001}, items 3--4). Each item is rated on a 4-point frequency scale (0 = \textit{Not at all}, 1 = \textit{Several days}, 2 = \textit{More than half the days}, 3 = \textit{Nearly every day}), yielding a subscore range of 0--6 per construct. The datasets collect these weekly, and each subscore serves as a separate prediction target in our experiments.}
\Description{Table showing the four PHQ-4 questionnaire items grouped into Anxiety (GAD-2) and Depression (PHQ-2) subscores.}
\label{tab:phq4}
\begin{tabular}{@{} p{0.22\textwidth} p{0.6\textwidth} @{}}
\toprule
\textbf{Subscore} & \textbf{Item: ``Over the last 2 weeks, how often have you been bothered by...''} \\
\midrule
\multirow{2}{=}{\textbf{Anxiety} (GAD-2, 0--6)}
  & 1.~Feeling nervous, anxious, or on edge \\
  & 2.~Not being able to stop or control worrying \\
\midrule
\multirow{2}{=}{\textbf{Depression} (PHQ-2, 0--6)}
  & 3.~Little interest or pleasure in doing things \\
  & 4.~Feeling down, depressed, or hopeless \\
\bottomrule
\end{tabular}
\vspace{-\baselineskip}
\end{table}

%% file: 5-results.tex
\section{Results}
\label{sec:results}

\begin{table*}[t]
\centering
\caption{Performance on the GLOBEM~\cite{xu2023globemdatasetmultiyeardatasets} dataset for anxiety and depression under the leave-one-subset-out evaluation protocol. Each entry reports test MAE $\pm$ SE. The best overall results are in \textbf{bold}, and the second best results are \underline{underlined}. \projectname{} consistently achieves the lowest error across all evaluation splits and tasks compared to both traditional and LLM-based baselines.}
\label{tab:globem_results}
\begin{tabular}{p{2.5cm}lcccc}
\toprule
\multirow{2}{*}{\textbf{Category}} & \multirow{2}{*}{\textbf{Method}} & \multicolumn{4}{c}{\textbf{Target Subset (MAE $\downarrow$)}} \\
\cmidrule(lr){3-6}
 & & \textbf{DS2} & \textbf{DS3} & \textbf{DS4} & \textbf{LOSO Avg} \\
\midrule
\multicolumn{6}{c}{\textit{Task: Anxiety Prediction}} \\
\midrule
\multirow{4}{=}{Trad. Feature-Based} & Linear Regression & 1.41$_{\pm\,0.03}$ & 1.30$_{\pm\,0.04}$ & 1.32$_{\pm\,0.03}$ & 1.35$_{\pm\,0.03}$ \\
 & SVR & 1.34$_{\pm\,0.03}$ & 1.38$_{\pm\,0.04}$ & 1.34$_{\pm\,0.03}$ & 1.35$_{\pm\,0.03}$ \\
 & XGBoost & 1.36$_{\pm\,0.03}$ & 1.39$_{\pm\,0.04}$ & 1.37$_{\pm\,0.03}$ & 1.38$_{\pm\,0.03}$ \\
 & MLP & 1.47$_{\pm\,0.03}$ & 1.30$_{\pm\,0.04}$ & 1.38$_{\pm\,0.04}$ & 1.38$_{\pm\,0.04}$ \\
\addlinespace
\multirow{3}{=}{SOTA Non-LLM Behavioral} & Ahmed et al.~\cite{ahmed_explainable_2025} & 1.34$_{\pm\,0.03}$ & 1.32$_{\pm\,0.03}$ & 1.31$_{\pm\,0.03}$ & 1.33$_{\pm\,0.03}$ \\
 & Borelli et al.~\cite{borelli_detection_2025} & 1.31$_{\pm\,0.03}$ & 1.25$_{\pm\,0.03}$ & \underline{1.28$_{\pm\,0.03}$} & \underline{1.28$_{\pm\,0.03}$} \\
 & ReOrder~\cite{xu_globem_2023} & 1.33$_{\pm\,0.03}$ & \underline{1.23$_{\pm\,0.03}$} & 1.32$_{\pm\,0.03}$ & 1.29$_{\pm\,0.03}$ \\
\addlinespace
\multirow{2}{=}{LLM-Based} & GPT-5.0~\cite{openai_gpt5_2026} & \underline{1.30$_{\pm\,0.08}$} & 1.36$_{\pm\,0.07}$ & 1.42$_{\pm\,0.07}$ & 1.37$_{\pm\,0.07}$ \\
 & Qwen3-4B~\cite{qwen3technicalreport} & 2.02$_{\pm\,0.03}$ & 2.33$_{\pm\,0.05}$ & 2.31$_{\pm\,0.04}$ & 2.22$_{\pm\,0.04}$ \\
\addlinespace
\textbf{Proposed} & \textbf{TimeSRL Qwen3-4B} & \textbf{1.25$_{\pm\,0.03}$} & \textbf{1.21$_{\pm\,0.04}$} & \textbf{1.26$_{\pm\,0.03}$} & \textbf{1.24$_{\pm\,0.03}$} \\
\midrule
\multicolumn{6}{c}{\textit{Task: Depression Prediction}} \\
\midrule
\multirow{4}{=}{Trad. Feature-Based} & Linear Regression & 1.39$_{\pm\,0.03}$ & 1.29$_{\pm\,0.04}$ & 1.35$_{\pm\,0.03}$ & 1.34$_{\pm\,0.03}$ \\
 & SVR & 1.33$_{\pm\,0.03}$ & 1.32$_{\pm\,0.04}$ & 1.37$_{\pm\,0.03}$ & 1.34$_{\pm\,0.03}$ \\
 & XGBoost & 1.35$_{\pm\,0.03}$ & 1.33$_{\pm\,0.04}$ & 1.37$_{\pm\,0.03}$ & 1.35$_{\pm\,0.03}$ \\
 & MLP & 1.34$_{\pm\,0.03}$ & 1.32$_{\pm\,0.04}$ & 1.35$_{\pm\,0.04}$ & 1.34$_{\pm\,0.03}$ \\
\addlinespace
\multirow{3}{=}{SOTA Non-LLM Behavioral} & Ahmed et al.~\cite{ahmed_explainable_2025} & 1.31$_{\pm\,0.03}$ & 1.34$_{\pm\,0.04}$ & 1.34$_{\pm\,0.03}$ & 1.33$_{\pm\,0.03}$ \\
 & Borelli et al.~\cite{borelli_detection_2025} & 1.25$_{\pm\,0.03}$ & \underline{1.26$_{\pm\,0.03}$} & 1.30$_{\pm\,0.03}$ & 1.27$_{\pm\,0.03}$ \\
 & ReOrder~\cite{xu_globem_2023} & \underline{1.21$_{\pm\,0.03}$} & 1.26$_{\pm\,0.04}$ & \underline{1.29$_{\pm\,0.03}$} & \underline{1.26$_{\pm\,0.03}$} \\
\addlinespace
\multirow{2}{=}{LLM-Based} & GPT-5.0~\cite{openai_gpt5_2026} & 1.28$_{\pm\,0.07}$ & 1.92$_{\pm\,0.08}$ & 1.83$_{\pm\,0.09}$ & 1.68$_{\pm\,0.08}$ \\
 & Qwen3-4B~\cite{qwen3technicalreport} & 2.79$_{\pm\,0.04}$ & 2.92$_{\pm\,0.06}$ & 2.95$_{\pm\,0.05}$ & 2.88$_{\pm\,0.05}$ \\
\addlinespace
\textbf{Proposed} & \textbf{TimeSRL Qwen3-4B} & \textbf{1.19$_{\pm\,0.03}$} & \textbf{1.22$_{\pm\,0.04}$} & \textbf{1.26$_{\pm\,0.03}$} & \textbf{1.22$_{\pm\,0.03}$} \\
\bottomrule
\end{tabular}
\end{table*}

We report the performance of \projectname{} on GLOBEM under the established cross-dataset evaluation protocol (Sec.~\ref{sub:results:overall_performance}), then evaluate its robustness on a second dataset (Sec.~\ref{sub:results:generalize_datasets}), examine generalizability across different LLM backbones (Sec.~\ref{sub:results:backbone_gain_interpretation}), test cross-study transfer under sensing pipeline variation (Sec.~\ref{sub:results:transfer}), conduct ablation studies to interpret the contribution of each steps (Sec.~\ref{sub:results:ablation}), and provide qualitative case study of the learned semantic abstractions (Sec.~\ref{sub:results:qualitative}).

\subsection{Strong Generalization Performance across Multiple Mental Health Prediction Tasks}
\label{sub:results:overall_performance}
Table~\ref{tab:globem_results} reports Leave-One-Dataset-Out (LOSO) results on GLOBEM to evaluate the robustness of \projectname{} under cross-cohort and cross-period distribution shift.
We measured model performance on two mental health prediction tasks: anxiety prediction and depression prediction.
Overall, \projectname{} achieves the best performance in both tasks consistently across all dataset splits (i.e., DS2, DS3, DS4 of GLOBEM). 

For anxiety, \projectname{} (LOSO MAE=1.24) outperforms the strongest prior non-LLM baseline (Borelli et al.~\cite{borelli_detection_2025}, MAE=1.28) by a relative improvement of 3.1\% ($p<0.05$).
The advantage over direct-prompting LLM baselines is substantially larger: compared to direct GPT-5.0 (MAE = 1.37), \projectname{} improves over a much larger and more powerful LLM baseline (GPT-5.0 is at least 100x bigger than our 4B model~\cite{qwen3technicalreport}) by 9.6\% ($p<0.001$); compared to direct Qwen3-4B (MAE = 2.22), our approach improve the same backbone LLM's performance by 44.1\% ($p<0.001$).

We observe similar performance on the second task on depression, \projectname{} achieves an LOSO MAE of 1.22, improving over the strongest prior non-LLM baseline (ReOrder~\cite{xu_globem_2023}) by 3.2\% ($p<0.001$). Similarly, the advantage compared to direct LLM prompting is more significant: \projectname{} outperforms the GPT-5.0 baseline (MAE= 1.68) by 27.4\% ($p<0.001$) and the Qwen3-4B baseline (MAE=2.88) by 57.6\% ($p<0.001$).

These improvements across multiple datasets in GLOBEM demonstrate the generalization capability of our proposed \projectname{} over the baseline approaches. It is noteworthy that GLOBEM includes subsets pre- and post-COVID outbreaks where behavioral distributions (DS3 and DS4) differ substantially from earlier data (DS2). Our results suggest that by extracting behavior semantics from the data, \projectname{} can handle such distribution shifts more robustly.

This directly supports the central motivation of \projectname{}. The benefit of our method does not come from simply replacing a conventional predictor with an LLM, nor from the size of LLM.
In fact, naive direct prompting performs substantially worse than task-specific traditional ML baselines,
and even a much stronger model such as GPT-5.0 still falls well short of \projectname{}. Instead, the gains emerge when the model is required to pass through an explicit semantic bottleneck and when that bottleneck is optimized for the final clinical objective.
Our experiment results \textbf{validate our key assumption of this paper}: under meaningful distribution shift, reasoning over outcome-aligned behavioral abstractions provides a more stable basis for inference than relying directly on brittle numerical regularities in the raw behavioral data.
For the rest of the result section, we further evaluate the robustness of \projectname{} across settings.

\begin{table*}[t]
\centering
\caption{Performance on the College Experience~\cite{nepal_capturing_2024} dataset for anxiety and depression under the leave-one-subset-out evaluation protocol. Each entry reports test MAE $\pm$ SE. The best overall results are in \textbf{bold}, and the second best results are \underline{underlined}. TimeSRL achieves top-tier performance on this dataset, significantly out-performing LLM methods and remains highly competitive with the strongest non-LLM methods on average.}
\label{tab:college_results}
\begin{tabular}{p{2.5cm}lcccc}
\toprule
\multirow{2}{*}{\textbf{Category}} & \multirow{2}{*}{\textbf{Method}} & \multicolumn{4}{c}{\textbf{Target Subset (MAE $\downarrow$)}} \\
\cmidrule(lr){3-6}
 & & \textbf{DS2} & \textbf{DS3} & \textbf{DS4} & \textbf{Avg} \\
\midrule
\multicolumn{6}{c}{\textit{Task: Anxiety Prediction}} \\
\midrule
\multirow{4}{=}{Trad. Feature-Based} & Linear Regression & 1.04$_{\pm\,0.02}$ & 1.15$_{\pm\,0.02}$ & 1.20$_{\pm\,0.02}$ & 1.13$_{\pm\,0.02}$ \\
 & SVR & 1.08$_{\pm\,0.02}$ & 1.23$_{\pm\,0.02}$ & 1.28$_{\pm\,0.03}$ & 1.20$_{\pm\,0.02}$ \\
 & XGBoost & 1.15$_{\pm\,0.02}$ & 1.19$_{\pm\,0.02}$ & 1.27$_{\pm\,0.03}$ & 1.20$_{\pm\,0.02}$ \\
 & MLP & 1.07$_{\pm\,0.02}$ & 1.22$_{\pm\,0.03}$ & 1.24$_{\pm\,0.02}$ & 1.18$_{\pm\,0.02}$ \\
\addlinespace
\multirow{3}{=}{SOTA Non-LLM Behavioral} & Ahmed et al.~\cite{ahmed_explainable_2025} & 1.08$_{\pm\,0.02}$ & 1.15$_{\pm\,0.02}$ & 1.18$_{\pm\,0.02}$ & 1.14$_{\pm\,0.02}$ \\
 & Borelli et al.~\cite{borelli_detection_2025} & 1.04$_{\pm\,0.02}$ & \underline{1.12$_{\pm\,0.02}$} & \underline{1.17$_{\pm\,0.02}$} & 1.11$_{\pm\,0.02}$ \\
 & ReOrder~\cite{xu_globem_2023} & \underline{1.02$_{\pm\,0.02}$} & \textbf{1.10$_{\pm\,0.02}$} & \textbf{1.15$_{\pm\,0.02}$} & \underline{1.09$_{\pm\,0.02}$} \\
\addlinespace
\multirow{2}{=}{LLM-Based} & GPT-5.0~\cite{openai_gpt5_2026} & 1.56$_{\pm\,0.08}$ & 1.72$_{\pm\,0.08}$ & 1.47$_{\pm\,0.08}$ & 1.58$_{\pm\,0.08}$ \\
 & Qwen3-4B~\cite{qwen3technicalreport} & 2.32$_{\pm\,0.03}$ & 2.51$_{\pm\,0.03}$ & 2.24$_{\pm\,0.03}$ & 2.36$_{\pm\,0.03}$ \\
\addlinespace
\textbf{Proposed} & \textbf{TimeSRL Qwen3-4B} & \textbf{1.00$_{\pm\,0.02}$} & \textbf{1.10$_{\pm\,0.02}$} & \textbf{1.15$_{\pm\,0.02}$} & \textbf{1.08$_{\pm\,0.02}$} \\
\midrule
\multicolumn{6}{c}{\textit{Task: Depression Prediction}} \\
\midrule
\multirow{4}{=}{Trad. Feature-Based} & Linear Regression & 1.07$_{\pm\,0.02}$ & 1.20$_{\pm\,0.02}$ & \textbf{1.24$_{\pm\,0.03}$} & 1.17$_{\pm\,0.02}$ \\
 & SVR & 1.11$_{\pm\,0.02}$ & 1.25$_{\pm\,0.02}$ & 1.41$_{\pm\,0.03}$ & 1.25$_{\pm\,0.02}$ \\
 & XGBoost & 1.18$_{\pm\,0.02}$ & 1.21$_{\pm\,0.02}$ & 1.34$_{\pm\,0.03}$ & 1.24$_{\pm\,0.03}$ \\
 & MLP & 1.06$_{\pm\,0.02}$ & 1.22$_{\pm\,0.02}$ & 1.34$_{\pm\,0.03}$ & 1.21$_{\pm\,0.03}$ \\
\addlinespace
\multirow{3}{=}{SOTA Non-LLM Behavioral} & Ahmed et al.~\cite{ahmed_explainable_2025} & 1.11$_{\pm\,0.02}$ & 1.17$_{\pm\,0.02}$ & 1.30$_{\pm\,0.03}$ & 1.19$_{\pm\,0.02}$ \\
 & Borelli et al.~\cite{borelli_detection_2025} & 1.07$_{\pm\,0.02}$ & 1.14$_{\pm\,0.02}$ & \underline{1.25$_{\pm\,0.03}$} & \underline{1.15$_{\pm\,0.02}$} \\
 & ReOrder~\cite{xu_globem_2023} & \textbf{1.00$_{\pm\,0.02}$} & \underline{1.09$_{\pm\,0.02}$} & 1.31$_{\pm\,0.03}$ & \textbf{1.13$_{\pm\,0.02}$} \\
\addlinespace
\multirow{2}{=}{LLM-Based} & GPT-5.0~\cite{openai_gpt5_2026} & 1.56$_{\pm\,0.08}$ & 1.55$_{\pm\,0.07}$ & 1.98$_{\pm\,0.09}$ & 1.70$_{\pm\,0.08}$ \\
 & Qwen3-4B~\cite{qwen3technicalreport} & 3.32$_{\pm\,0.04}$ & 3.39$_{\pm\,0.04}$ & 3.06$_{\pm\,0.04}$ & 3.26$_{\pm\,0.04}$ \\
\addlinespace
\textbf{Proposed} & \textbf{TimeSRL Qwen3-4B} & \underline{1.05$_{\pm\,0.02}$} & \textbf{0.95$_{\pm\,0.03}$} & 1.38$_{\pm\,0.04}$ & \textbf{1.13$_{\pm\,0.03}$} \\
\bottomrule
\end{tabular}
\end{table*}

\begin{figure}[t]
    \centering
    \includegraphics[width=0.95\linewidth]{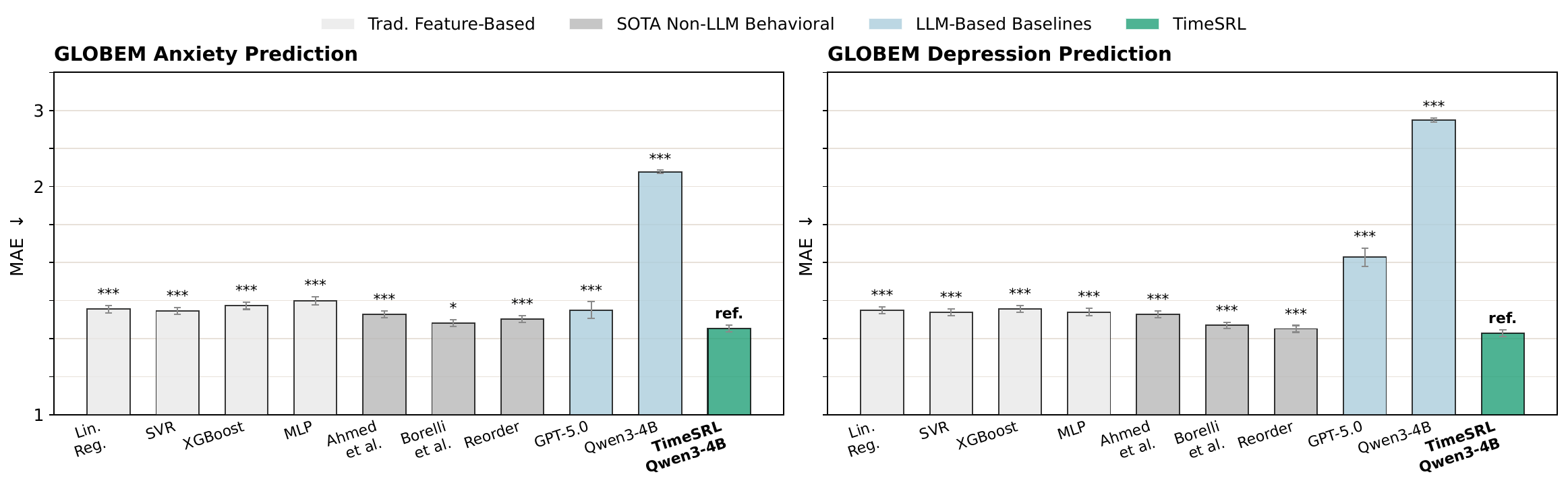}
    \vspace{0.5em}
    \includegraphics[width=0.95\linewidth]{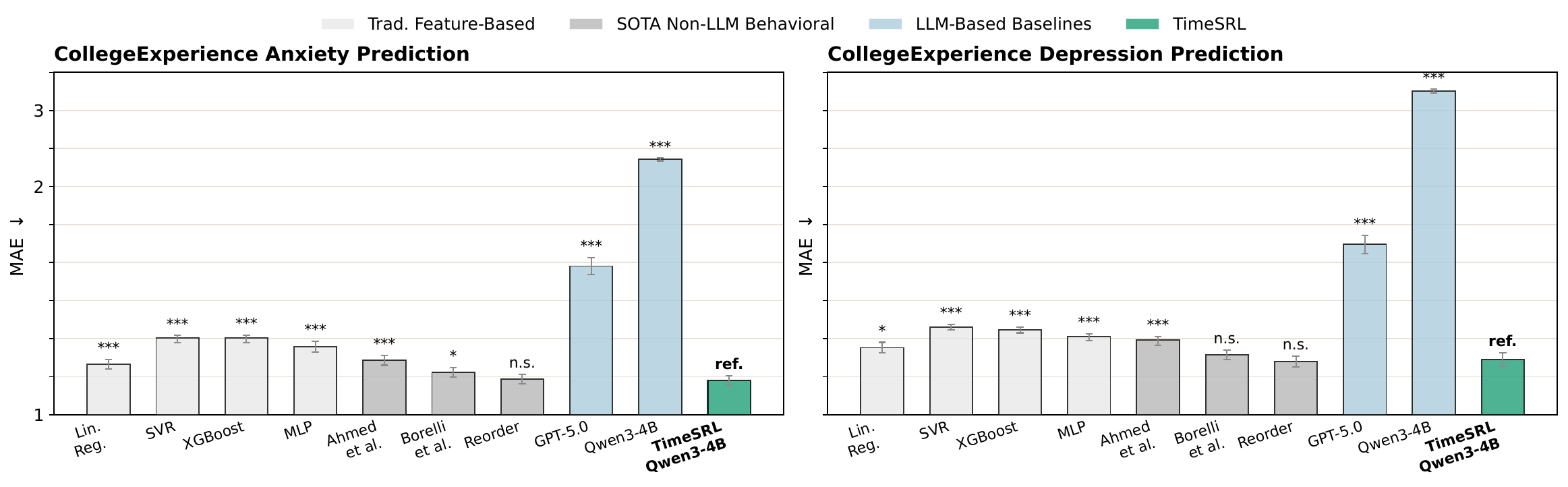}
    \caption{\textbf{LOSO MAE on GLOBEM and College Experience anxiety and depression prediction.} Bars denote mean MAE and error bars denote 95\% percentile bootstrap SE. Star annotations report paired bootstrap significance tests versus \projectname{}, indicating significantly higher MAE than \projectname{}, and significance levels are marked as $^{*}p<0.05$, $^{**}p<0.01$, and $^{***}p<0.001$. Across both datasets and tasks, TimeSRL maintains top-tier performance, significantly outperforming all other LLM baselines and remaining highly competitive with or surpassing the best non-LLM behavioral models.}
    \label{fig:bootstrap}
\end{figure}

\subsection{\projectname{} Generalizes across Multiple Datasets} 
\label{sub:results:generalize_datasets}
Table~\ref{tab:college_results} reports LOSO results on College Experience to test whether the same framework remains effective in a second longitudinal sensing setting. \projectname{} again outperforms all LLM baselines and remains competitive with the strongest non-LLM methods across both tasks, though the margins are more modest than on GLOBEM.

Unlike GLOBEM, College Experience was not designed as a cross-cohort generalization benchmark. Although the curated splits are subject-disjoint and time-disjoint, they are drawn from the same underlying college cohort and do not reflect a major external shift such as the onset of COVID. With less pressure to handle sharply different behavioral regimes, there is also less room for an abstraction mechanism to separate itself from strong feature-based baselines.

Even in this milder setting, \projectname{} retains its advantage. For anxiety, it improves over the strongest baseline (ReOrder, MAE = 1.09) to 1.08. For depression, it achieves the best average MAE of 1.13, matching ReOrder on average while achieving the lowest error on one of the three held-out splits (DS3, MAE = 0.95). Meanwhile, its advantage over the LLM baselines remains substantial: for anxiety, \projectname{} outperforms GPT-5.0 (MAE = 1.58) by 31.6\% and vanilla Qwen3-4B (MAE = 2.36) by 54.2\%; for depression, it outperforms GPT-5.0 (MAE = 1.70) by 33.5\% and vanilla Qwen3-4B (MAE = 3.26) by 65.3\%.

As Figure~\ref{fig:bootstrap} summarizes, \projectname{} is the best or top-tier method across both tasks and both datasets, with paired bootstrap tests confirming statistical significance against all LLM baselines ($p\text{s}<0.001$).
Together, these results confirm that the proposed bottleneck generalizes beyond a single benchmark dataset, while also indicating that its value is largest when the inference problem involves a stronger distribution shift.

\begin{figure*}[t]
  \centering
  \includegraphics[width=\textwidth]{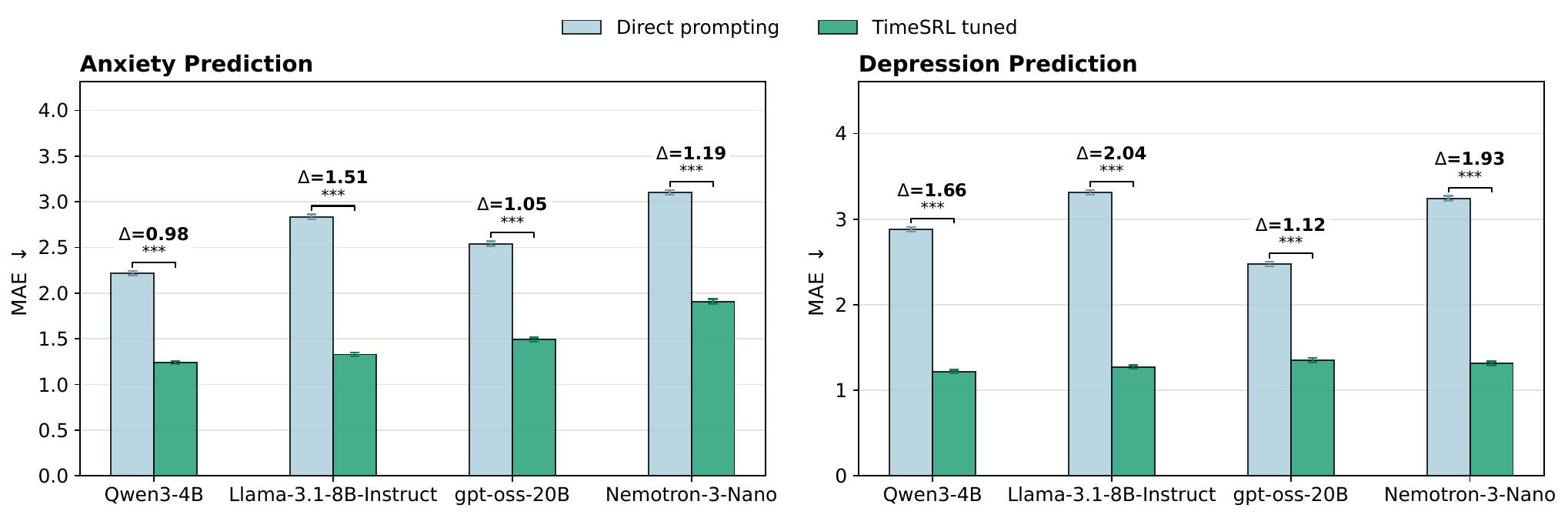}
  \caption{\textbf{MAE reduction from \projectname{} tuning across four LLM backbones on GLOBEM (LOSO).} Bars compare direct prompting against the \projectname{}-tuned variant for anxiety and depression; error bars denote standard error. \projectname{} consistently improves every backbone, with relative MAE reductions of 38.4--61.6\% across all backbone--task combinations.}
  \label{fig:backbone_tuning_gain}
  \vspace{-\baselineskip}
\end{figure*}

\subsection{\projectname{} Generalizes across Multiple LLM Backbones}
\label{sub:results:backbone_gain_interpretation}
A benefit tied to a single LLM backbone would limit the practical value of the framework. To test method portability, we apply the same \projectname{} tuning procedure to three additional backbones beyond Qwen3-4B: Llama-3.1-8B-Instruct, GPT-oss-20B, and Nemotron-3-Nano. Figure~\ref{fig:backbone_tuning_gain} reports the results under the LOSO protocol on GLOBEM. Across all tested models, the tuned version consistently outperforms direct prompting on both anxiety and depression prediction, with relative MAE reductions ranging from 38.4\% to 61.6\%.

For Qwen3-4B, the main backbone used throughout this paper, \projectname{} reduces MAE from 2.22 to 1.24 on anxiety (44.1\% reduction) and from 2.88 to 1.22 on depression (57.6\%). Llama-3.1-8B-Instruct shows the largest absolute improvement, with MAE dropping from 2.83 to 1.33 on anxiety (53.1\%) and from 3.31 to 1.27 on depression (61.6\%). GPT-oss-20B improves from 2.54 to 1.49 on anxiety (41.4\%) and from 2.47 to 1.35 on depression (45.3\%). Nemotron-3-Nano, starting from the weakest direct-prompting baseline (MAE 3.10 and 3.24), improves to 1.91 and 1.31, corresponding to reductions of 38.4\% and 59.5\% on anxiety and depression respectively. All improvements showed strong statistical significance (ps<0.001). These gains are not marginal refinements; they indicate that the proposed training procedure substantially improves how each backbone abstracts and reasons over the behavioral trajectory relative to its starting performance. While every backbone did not benefit equally, it demonstrates that the mechanism is not a one-model artifact. This strengthens the broader claim that the value of \projectname{} lies in the training framework and the learned semantic bottleneck, not in a particular backbone choice.

\begin{figure*}[t]
  \centering
  \includegraphics[width=\textwidth]{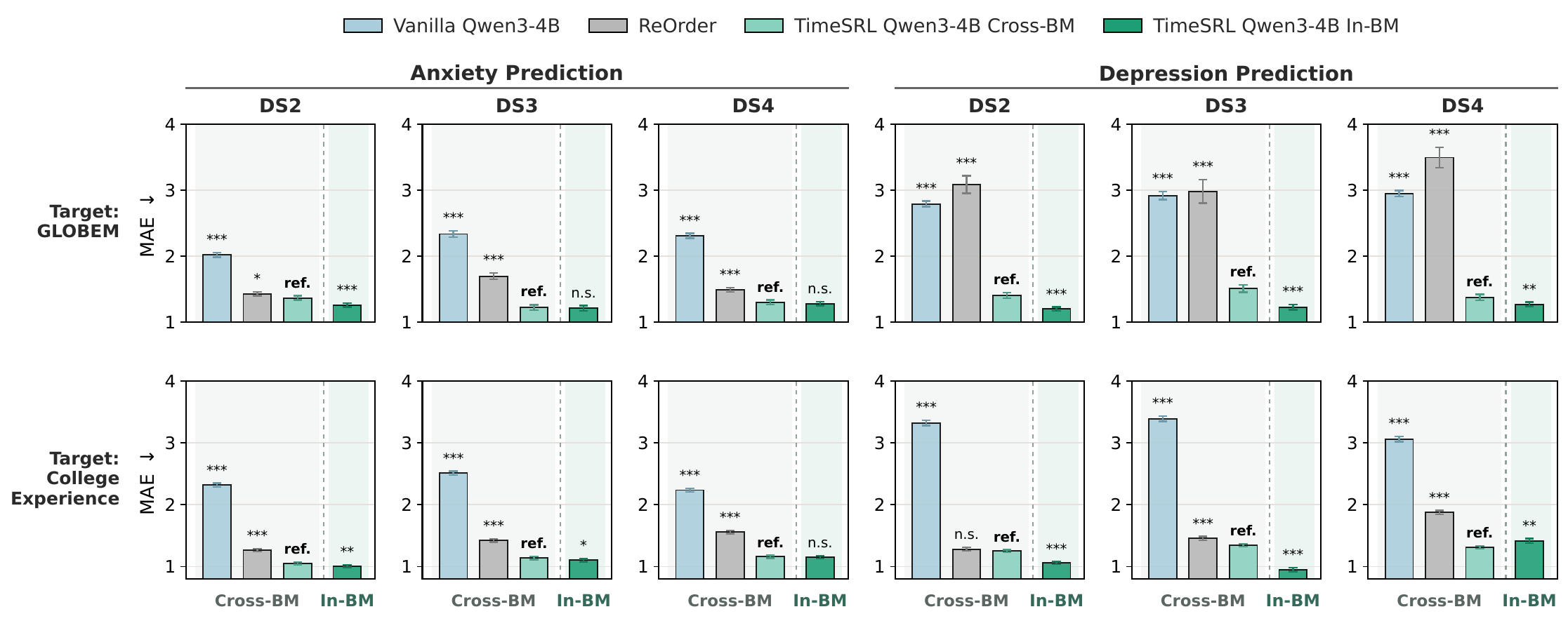}
  \caption{ \textbf{Cross-benchmark (Cross-BM) transfer results on GLOBEM and College Experience.} Each panel evaluates transfer in one target benchmark split after training on the other benchmark; the rightmost bar
  shows the within-benchmark (In-BM) TimeSRL reference for the same target study. Stars denote statistical significance vs.\ the Cross-BM reference (paired bootstrap; {}$p<0.05$, {}$p<0.01$, {}$p<0.001$; n.s.\ =
   not significant). TimeSRL shows strong promise for cross-benchmark transfer, consistently outperforming both vanilla prompting and the strongest prior baseline overall (ReOrder), and
  sometimes matching or outperforming the In-BM.}
  \label{fig:cross_dataset_transfer}
\end{figure*}

\subsection{\projectname{} Shows Promises to Transfer Insights across Benchmarks even without Finetuning}
\label{sub:results:transfer}
The within-benchmark evaluations in Sections~\ref{sub:results:overall_performance} and~\ref{sub:results:generalize_datasets} test generalization under population and temporal shift, but the sensing infrastructure remains the same within each benchmark. If \projectname{} follows our key assumption as in Sections~\ref{sub:results:overall_performance}, the insights learned from one benchmark's task should be able to \textit{directly} generalize to the same task in another benchmark without any fine-tuning.

We evaluate this more challenging setting: we train \projectname{} to do anxiety or depression prediction on one benchmark and evaluate it on the other (\ie training on College Experience and evaluating on GLOBEM, and vice versa), comparing the transferred model against vanilla Qwen3-4B prompting, ReOrder~\cite{xu_globem_2023}, the within-benchmark \projectname{} model (\ie training and evaluating on the same benchmark).
Because GLOBEM and College Experience use different collection devices and processing pipelines, even identically named features may exhibit distributional drift unrelated to behavioral differences. Notably, the LLM-based approaches are able to consume each study's full native feature set without modification, whereas ReOrder can only be trained on a shared subset of proxy features that could be aligned across the two studies. This cross-benchmark protocol therefore tests whether the semantic abstractions learned by \projectname{} are robust to feature-space variation introduced by the sensing and data collection pipeline itself.

Cross-benchmark \projectname{} outperforms both vanilla LLM prompting and ReOrder across every task and every data split (Figure~\ref{fig:cross_dataset_transfer}).
On the GLOBEM target (\ie models trained on College Experience), the transferred model achieves an average MAE of 1.29 on anxiety and 1.43 on depression, compared with 2.22 (41.9\% improvement) and 2.88 (50.3\% improvement) for vanilla LLM prompting, and 1.54 (16.2\% improvement) and 3.19 (55.2\% improvement) for ReOrder.
On the College Experience target (\ie models trained on GLOBEM), the improvements are also consistent: \projectname{} outperforms vanilla LLM prompting by 60.1\% and ReOrder by 15.6\% on depression, as well as 52.5\% and 21.1\% on anxiety, respectively. All improvements are statistically significant ($p$s$<0.001$).
The gains are especially pronounced in depression when GLOBEM is the target benchmark. ReOrder's manually aligned feature set fails to transfer effectively. Its transferred MAE (3.19) is actually higher than vanilla LLM prompting (2.88), while \projectname{}'s learned semantic abstraction reduces MAE by 50.3\% relative to vanilla LLM prompting and 55.2\% relative to ReOrder.
In fact, the results of the cross-benchmark \projectname{} are even comparable to those of the within-benchmark \projectname{} (e.g., 1.24 vs.\ 1.29 on anxiety with GLOBEM as target, 1.08 vs.\ 1.12 with College Experience as target).

These results indicate that the behavioral structure captured by \projectname{} is reusable rather than benchmark-specific or feature-specific. Unlike ReOrder, which relies on a fixed set of hand-aligned behavior features, \projectname{} learns to abstract each benchmark's native feature space into semantic descriptions of behavioral patterns. This abstraction is expressive enough to transfer across heterogeneous sensing conditions and behavior features without any target-domain fine-tuning.
For ubiquitous sensing applications where labeled target data may be scarce or unavailable, \projectname{} offers a practical and effective cross-benchmark starting point that already surpasses baselines.

\definecolor{degradeL}{HTML}{D4896E}
\definecolor{degradeH}{HTML}{8B0000}
\newcommand{\pmark}[2]{{\scriptsize\textcolor{#1}{(#2\%)}}}

\begin{table*}[t]
\caption{%
  Ablation study of TimeSRL components.
  We vary two design axes---prompting strategy (direct vs.\ 2-Stage)
  and tuning (none vs.\ RL)---across Qwen3-4B and GPT-5.0 base models.
  MAE~($\downarrow$) is reported with relative change (\%) from TimeSRL;
  \textcolor{degradeL}{light red} denotes $<$50\% degradation,
  \textcolor{degradeH}{dark red} denotes $\geq$50\%.
  All differences are statistically significant
  ($p\text{s} < 0.001$, paired bootstrap, 5\,000 resamples). Combination of RL tuning and 2-Stage prompting outperforms both alone.
}
\label{tab:ablation}
\centering
\small
\begin{tabular}{@{}l l  r@{\,}l  r@{\,}l  r@{\,}l  r@{\,}l@{}}
\toprule
& &
\multicolumn{4}{c}{\textbf{GLOBEM}} &
\multicolumn{4}{c}{\textbf{CollegeExperience}} \\
\cmidrule(lr){3-6} \cmidrule(lr){7-10}
\textbf{Base Model} &
\textbf{Configuration} &
\multicolumn{2}{c}{Anxiety} &
\multicolumn{2}{c}{Depression} &
\multicolumn{2}{c}{Anxiety} &
\multicolumn{2}{c}{Depression} \\
\midrule
\multirow{4}{*}{Qwen3-4B}
 & Direct prompting      & 2.19 & \pmark{degradeH}{+76.6}  & 2.87 & \pmark{degradeH}{+135.2} & 2.36 & \pmark{degradeH}{+118.5} & 3.25 & \pmark{degradeH}{+187.6} \\
 & 2-Stage                & 3.42 & \pmark{degradeH}{+175.8} & 4.20 & \pmark{degradeH}{+244.3} & 3.94 & \pmark{degradeH}{+264.8} & 4.58 & \pmark{degradeH}{+305.3} \\
 & Direct prompting + RL  & 1.34 & \pmark{degradeL}{+8.1}   & 1.34 & \pmark{degradeL}{+9.8}   & 1.13 & \pmark{degradeL}{+4.6}   & 1.25 & \pmark{degradeL}{+10.6}  \\
\cmidrule(l){2-10}
 & 2-Stage + RL \textbf{(TimeSRL)} & \textbf{1.24} & & \textbf{1.22} & & \textbf{1.08} & & \textbf{1.13} & \\
\midrule
\multirow{2}{*}{GPT-5.0}
 & Direct prompting       & 1.37 & \pmark{degradeL}{+9.8}   & 1.63 & \pmark{degradeL}{+33.6}  & 1.58 & \pmark{degradeL}{+46.3}  & 1.70 & \pmark{degradeH}{+50.4}  \\
 & 2-Stage                & 1.66 & \pmark{degradeL}{+33.9}  & 1.92 & \pmark{degradeH}{+57.4}  & 1.82 & \pmark{degradeH}{+68.5}  & 2.22 & \pmark{degradeH}{+96.5}  \\
\bottomrule
\end{tabular}
\end{table*}

\subsection{Insights from Ablation Studies: Combination of RL and 2-Stage Outperforms Both Alone}
\label{sub:results:ablation}
\projectname{} integrated two essential components: two-stage reasoning and abstraction process, and RL-based finetuning.
To determine which components drive the observed gains, we compare four configurations: (1)~\emph{direct one-stage prediction}, where the LLM predicts the score from behavioral logs in a single step; (2)~\emph{two-stage prompting without RL}, which preserves the abstract-then-predict decomposition but uses no GRPO optimization; (3)~\emph{direct one-stage prediction with RL}, which applies GRPO without an intermediate abstraction step; and (4)~the full \projectname{} pipeline. We evaluate all four configurations using Qwen3-4B and the configurations without RL using GPT-5.0 (a proprietary model), on both benchmarks. Table~\ref{tab:ablation} reports the results.

We found that a two-stage decomposition by itself is not sufficient, and can actually be \textit{harmful}. Across all four evaluation settings, untuned two-stage prompting performs worse than direct prediction. For GPT-5.0 on GLOBEM, MAE worsens from 1.37 to 1.66 on anxiety (21.2\%) and from 1.63 to 1.92 on depression (17.8\%). The same pattern holds on College Experience, where GPT-5.0 degrades from 1.58 to 1.82 on anxiety (15.2\%) and from 1.70 to 2.22 on depression (30.6\%). Qwen3-4B shows an even larger drop. All differences show statistical significance (ps<0.001). Inserting a summary step without learning how that summary should preserve evidence does not help the model reason over behavioral trajectories.

Meanwhile, RL alone already contributes a good portion of the performance improvement. Relative to untuned Qwen3-4B, direct prediction RL dramatically reduces error on every benchmark. On GLOBEM, MAE drops 38.8\% for anxiety and from 53.3\% for depression. On College Experience, it drops from 52.1\% for anxiety and from 61.5\% for depression (ps<0.001). These gains show that outcome-aligned optimization helps the model extract more useful predictive structure from longitudinal behavioral data, even without an explicit intermediate abstraction. However, RL alone did not outperform the best non-LLM ML baselines.

More importantly, the full method achieves the best results, indicating that the integration of the two-stage decomposition and RL can amplify the synergies between these two components. Compared with direct prediction RL, \projectname{} further reduces MAE 7.5\% on anxiety and 9.0\% on depression in GLOBEM, and 4.4\% on anxiety and 9.6\% on depression in College Experience. The strongest results come from combining optimization with an explicit semantic bottleneck, not from either ingredient in isolation.

The ablation also argues against a pure model-scale explanation. Direct GPT-5.0 is much stronger than untuned Qwen3-4B, yet it underperformed the finetuned Qwen3-4B using \projectname{} on every setting in Table~\ref{tab:ablation}. A larger and generally more powerful model can reason over the input more effectively, but it does not automatically learn the kind of intermediate representation that is most effective and robust under distribution shift. These results support our hypothesis that \emph{semantic abstraction is effective only when the abstraction itself is optimized towards the predictive target}.

\input{figs/qualitative_example}

\subsection{Qualitative Case Study: How the Semantic Bottleneck Shapes Intermediate Summaries}
\label{sub:results:qualitative}
To build intuition for \emph{how} the semantic bottleneck changes model behavior, we present an illustrative case study (Fig.~\ref{fig:qual_example_modern}) comparing the intermediate summaries produced by untuned two-stage baselines (GPT-5.0 and Qwen3-4B) against the RL-tuned \projectname{} Qwen3-4B model on the same 14-day behavioral window (additional examples are provided in Appendix~\ref{app:more_qual_examples}). The goal of this section is not to establish a general property of the learned summaries, but to make concrete one qualitative difference we observed when inspecting outputs: in this example, both untuned baselines substantially overpredict anxiety while \projectname{} recovers the correct mild score, and the underlying summaries differ in ways that plausibly explain the gap.

The \projectname{} semantic abstracted summary produced more text than the untuned versions (approx. 1500 vs 500 words, example shows redacted text due to space). However, the key difference is not simply that \projectname{} produces more text. Rather, it organizes the behavioral trajectory into interpretable dimensions, separates stable patterns from localized irregularities, and avoids collapsing the entire window into a uniformly high-risk narrative. In contrast, the untuned baselines compress the window around a small number of salient concerning events and then propagate that framing into inflated downstream scores.

In this example, we observe a pattern that may reflect a broader failure mode of untuned two-stage pipelines: the intermediate summary appears \emph{selectively compressed} around the most salient negative events in the window. In this case, both GPT-5.0 and untuned Qwen3-4B anchor heavily on severe but localized irregularities, especially the Day 9 sleep nadir, and reinterpret broader contextual signals such as high home stay or sparse social activity as persistent evidence of anxiety. Once the summary is framed in this way, the second-stage reasoner inherits that framing and produces an inflated score.

\projectname{} changes this behavior in two ways. First, it imposes a more explicit structure over the behavioral trajectory, separating dimensions such as sleep, activity, mobility, and phone use instead of blending them into a single risk narrative. Second, it preserves temporal calibration by distinguishing isolated disruptions from sustained patterns. In the example, the RL-tuned summary still captures the most concerning day (day 9 sleep), but it also retains evidence of stability, recovery, and non-monotonic variation across the remainder of the window. This yields a representation that is more faithful to the full trajectory and more useful for downstream scoring.

While a few examples cannot establish a general property, this case is consistent with the hypothesis that the benefit of \projectname{} is not generic summarization quality alone, but rather the RL objective's pressure toward summaries that preserve decision-relevant behavioral structure. A systematic characterization of summary properties across the full test set should be explored in future work.

\subsection{Summary of Key Findings}
Taken together, the experiments support the following main findings:
\begin{s_enumerate}
\item \projectname{} achieves state-of-the-art generalization on GLOBEM benchmark, outperforming all non-LLM and LLM baselines with statistical significance across both anxiety and depression under a rigorous cross-cohort evaluation protocol (Sec.~\ref{sub:results:overall_performance}). 
\item Similar gains persist on another College Experience benchmark (Sec.~\ref{sub:results:generalize_datasets}) and generalize across four different LLM backbones (Sec.~\ref{sub:results:backbone_gain_interpretation}), confirming that the advantage stems from the training framework rather than a particular benchmark or model choice.
\item Cross-benchmark transfer experiments show that \projectname{} has the potential to transfer its learned abstractions across benchmarks with different sensing pipelines (Sec.~\ref{sub:results:transfer}), demonstrating robustness to feature-space drift beyond population and temporal shift.
\item Ablation studies reveal that neither RL optimization nor two-stage decomposition is sufficient alone: untuned two-stage prompting actively hurts performance, RL alone improves substantially but does not surpass the best non-LLM baselines, and only the combination of both components achieves state-of-the-art results (Sec.~\ref{sub:results:ablation}). 
\item Qualitative case study suggests that \projectname{} succeeds by learning intermediate summaries that are \emph{decision-useful}, preserving balanced, temporally calibrated representations of behavioral trajectories rather than collapsing them into risk-focused narratives (Sec.~\ref{sub:results:qualitative}).
\end{s_enumerate}

%% file: figs/qualitative_example.tex
\begin{figure*}[t]
\centering
\small
\setlength{\tabcolsep}{6pt}
\renewcommand{\arraystretch}{1.12}

\begin{tcolorbox}[
    enhanced,
    colback=white,
    colframe=black!15,
    boxrule=0.5pt,
    arc=2mm,
    left=1.2mm,right=1.2mm,top=1.2mm,bottom=1.2mm
]
\begin{tabularx}{\textwidth}{@{}p{0.18\textwidth}X@{}}
\toprule
\textbf{Example sample} & \texttt{GLOBEM INS-W\_603\_2020-05-17} \\
\textbf{Gold score} & \textbf{1} \\
\textbf{Predictions} & GPT-5.0: 4 | Vanilla Qwen3-4B: 6 | \textbf{TimeSRL Qwen3-4B: 1} \\
\bottomrule
\end{tabularx}
\end{tcolorbox}

\vspace{0.5em}

\begin{minipage}[t]{0.32\textwidth}
\begin{tcolorbox}[
    enhanced,
    colback=baselinebg,
    colframe=baselinegray!45,
    boxrule=0.8pt,
    arc=2mm,
    left=1.2mm,right=1.2mm,top=1.0mm,bottom=1.0mm,
    title=\textbf{GPT-5.0 (2-Stage baseline)},
    coltitle=black,
    fonttitle=\small\bfseries
]
\textbf{Stage-1 summary excerpt}

\vspace{0.25em}
\footnotesize
Predominantly homebound across the 14 days, with rare bursts of mobility and novelty. Sleep is highly volatile, including several short nights and one severe nadir. Phone checking spikes on short-sleep days, with frequent near-midnight use. In-person social density is minimal. \textcolor{warnred}{These patterns suggest recurrent hyperarousal, avoidance, and social withdrawal.}

\vspace{0.5em}
\begin{tcolorbox}[
    enhanced,
    colback=warnredbg,
    colframe=warnred!35,
    boxrule=0.5pt,
    arc=1.5mm,
    left=0.8mm,right=0.8mm,top=0.6mm,bottom=0.6mm
]
\footnotesize
\textbf{Compression pattern.} Concern-heavy cues dominate the summary, while stabilizing or recovering periods are only weakly represented.
\end{tcolorbox}

\vspace{0.35em}
\textbf{Stage-2 output}

\vspace{0.2em}
\footnotesize
\texttt{<think>} Multiple hyperarousal days, sustained homebound behavior, and minimal social contact indicate recurrent anxiety. Brief improvements are outweighed by the broader pattern. \texttt{</think>}

\texttt{score: 4}
\end{tcolorbox}
\end{minipage}\hfill
\begin{minipage}[t]{0.27\textwidth}
\begin{tcolorbox}[
    enhanced,
    colback=baselinebg,
    colframe=baselinegray!45,
    boxrule=0.8pt,
    arc=2mm,
    left=1.2mm,right=1.2mm,top=1.0mm,bottom=1.0mm,
    title=\textbf{Qwen3-4B (2-Stage baseline)},
    coltitle=black,
    fonttitle=\small\bfseries
]
\textbf{Stage-1 summary excerpt}

\vspace{0.25em}
\footnotesize
Time at home is consistently high except for a sharp drop on Day 7. Sleep duration is variable, with a concerning 2.3-hour night on Day 9. Mobility remains limited, phone use is high, and social density is minimal. \textcolor{warnred}{Overall, the pattern reflects isolation, disrupted sleep, and inconsistent activity, consistent with anxiety symptoms.}

\vspace{0.5em}
\begin{tcolorbox}[
    enhanced,
    colback=warnredbg,
    colframe=warnred!35,
    boxrule=0.5pt,
    arc=1.5mm,
    left=0.8mm,right=0.8mm,top=0.6mm,bottom=0.6mm
]
\footnotesize
\textbf{Compression pattern.} The trajectory is condensed into a single severity-oriented interpretation with limited temporal nuance.
\end{tcolorbox}

\vspace{0.35em}
\textbf{Stage-2 output}

\vspace{0.2em}
\footnotesize
\texttt{<think>} Sharp behavioral changes, poor sleep, and social isolation suggest a more severe pattern. \texttt{</think>}

\texttt{score: 6}
\end{tcolorbox}
\end{minipage}\hfill
\begin{minipage}[t]{0.40\textwidth}
\begin{tcolorbox}[
    enhanced,
    colback=timesrlbluebg,
    colframe=timesrlblue!65,
    boxrule=1.0pt,
    arc=2mm,
    left=1.2mm,right=1.2mm,top=1.0mm,bottom=1.0mm,
    title=\textbf{TimeSRL Qwen3-4B},
    coltitle=black,
    fonttitle=\small\bfseries
]
\textbf{Stage-1 summary excerpt (RL-tuned)}

\vspace{0.25em}
\scriptsize
\textbf{Overall trajectory:} Mixed stability with localized volatility across home stay, sleep, activity, communication, and phone use.

\textbf{Home stay:} Generally high, with notable drops on D7 and D11, then recovery near baseline by D14.

\textbf{Sleep:} Efficiency is mostly stable, while duration is volatile. D9 is a clear outlier, but longer nights appear on D7 and D14.

\textbf{Activity:} Short-lived peaks and dips suggest temporary fluctuations rather than sustained deterioration.

\textbf{Communication / phone use:} Unlocks and calls spike around D9, but other days are comparatively moderate.

\textbf{Volatility summary:} \textcolor{goodgreen}{D9 is the most concerning day, while several other days appear comparatively stable or partially normalized.}

\vspace{0.4em}
\emph{Excerpt shown for readability. The full output also includes additional dimensions such as unique places and green-space exposure.}

\vspace{0.5em}
\begin{tcolorbox}[
    enhanced,
    colback=goodgreenbg,
    colframe=goodgreen!35,
    boxrule=0.5pt,
    arc=1.5mm,
    left=0.8mm,right=0.8mm,top=0.6mm,bottom=0.6mm
]
\scriptsize
\textbf{Preservation pattern.} Concerning signals are retained, but they are explicitly localized in time and balanced with evidence of stability and recovery.
\end{tcolorbox}

\vspace{0.35em}
\textbf{Stage-2 output}

\vspace{0.2em}
\scriptsize
\texttt{<think>} The pattern shows isolated disruptions rather than persistent high-severity anxiety. Overall this is more consistent with mild symptoms. \texttt{</think>}

\texttt{score: 1}
\end{tcolorbox}
\end{minipage}

\vspace{0.4em}

\caption{\textbf{Qualitative comparison of intermediate summaries on a 14-day window.} Model-generated summaries and predictions are presented alongside highlighted framing sentences (colored text) and annotated compression/preservation patterns (shaded boxes). Both untuned two-stage baselines (GPT-5.0, Qwen3-4B) compress the trajectory into a predominantly concern-heavy narrative, emphasizing salient irregularities while under representing stable or recovering periods. TimeSRL instead produces a more structured and temporally localized summary. Although only an excerpt is shown for readability, the full RL-tuned output preserves both concerning and stabilizing evidence, which leads to a calibrated prediction matching the gold label.}
\label{fig:qual_example_modern}
\end{figure*}

%% file: 6-discussion.tex
\section{Discussion}
\label{sec:discussion}
The results in the previous section establish that an RL-optimized semantic bottleneck improves generalization for behavioral health prediction. In reflecting on these findings, three observations emerge that extend beyond the specific evaluation setting. First, the structural conditions that make the semantic bottleneck effective, including distribution shift, available end-task labels, and predictive signal residing in higher-level temporal patterns, are not unique to behavioral health, which raises the question of how broadly the framework may apply (Sec.~\ref{sub:discussion:scalability}). Second, the fact that the intermediate representation is expressed in natural language, rather than as a learned embedding, introduces interpretability properties that may be practically relevant in clinical deployment contexts (Sec.~\ref{sub:discussion:interpretability}). Third, the current instantiation operates within specific boundaries highlighted in limitations (\eg a particular model scale, prompt format, output range, and training regime) that naturally scope where the current findings apply and where further investigation is needed (Sec.~\ref{sub:discussion:limitations}).

\subsection{Scalability of the Semantic Bottleneck beyond Behavioral Health}
\label{sub:discussion:scalability}
Although we instantiate \projectname{} on mental-health prediction from passive sensing, the underlying framework (\ie an RL-optimized semantic bottleneck between raw time series and downstream inference) is not specific to this domain. Any setting where (1)~multivariate temporal data must be interpreted under distribution shift, (2)~validated end-task labels exist but gold intermediate annotations do not, and (3)~the predictive signal resides in higher-level temporal patterns rather than point-wise numerical values satisfies the same structural prerequisites. These conditions are common across time-series reasoning tasks in ubiquitous computing and beyond.

Consider clinical deterioration prediction from continuous physiological monitoring, where heart rate, blood pressure, and respiratory signals are collected across hospital stays that differ by patient demographics, sensor hardware, and care protocols. A semantic bottleneck could abstract these trajectories into descriptions of hemodynamic trends, autonomic stability, and recovery dynamics, providing a representation that transfers more naturally across institutions than raw vital-sign magnitudes. Similarly, in energy forecasting from building sensor networks, daily load profiles vary across climates, building types, and occupant populations; abstracting consumption trajectories into semantic patterns of occupancy-driven demand, diurnal cycling, and weather-sensitive load components could yield representations that generalize across deployment sites. Other natural candidates include activity recognition from wearable accelerometry, urban mobility modeling, and environmental monitoring, all domains where temporal structure is rich, distribution shift is pervasive, and end-task labels are available while intermediate semantic annotations are not.

More broadly, the semantic bottleneck can be viewed as a form of self-supervised representation learning that operates in natural language rather than in a learned embedding space. The RL objective acts as a task-aligned shaping signal: it does not require the practitioner to specify what the summary should contain, but instead lets the optimization discover which semantic properties are decision-useful for the target task. This makes the approach particularly attractive for domains where expert intuition about what constitutes a ``good'' intermediate representation is limited.

\subsection{Interpretability and Clinical Utility of the Semantic Bottleneck}
\label{sub:discussion:interpretability}
Building on the natural-language framing of the intermediate representation, an unusual situation arises in predictive modeling: the representation through which the prediction must pass is, by construction, human-readable. While the primary motivation for the semantic bottleneck is improved generalization under distribution shift, this interpretability opens practical opportunities for clinical workflows.

In current clinical practice, passive sensing models are largely opaque: a predicted symptom score arrives without explanation, and the clinician must decide whether to trust it based on the score alone~\cite{cai_hello_2019}. The \projectname{} summary offers a richer interface. A clinician reviewing the output can inspect the behavioral narrative---sleep fragmentation patterns, mobility trends, social engagement changes---and assess whether the summary aligns with their clinical knowledge of the patient. When the summary highlights a pattern that the clinician recognizes (e.g., progressive social withdrawal following a sleep disruption), the prediction becomes more actionable. When the summary contains implausible claims, the clinician has a concrete basis for discounting the prediction rather than simply ignoring an unexplained number.

This transparency also supports a natural human-in-the-loop workflow. Because the summary is the sole input to Stage~2, a domain expert could in principle edit or annotate the intermediate summary before the prediction is made, correcting mischaracterizations or adding contextual knowledge that the sensing data alone cannot capture. While we do not evaluate such interactive use in this work, the architecture is structurally compatible with it, unlike post hoc explanation methods that produce rationales disconnected from the model's actual decision process.

It is important to note, however, that interpretability of the intermediate summary does not guarantee faithfulness. The summary is shaped by RL to support accurate predictions, not to provide a clinically complete behavioral assessment. It may emphasize patterns that are statistically predictive but clinically non-obvious, or omit contextual factors that a clinician would consider important. The summaries should therefore be understood as \emph{decision-useful abstractions optimized for a specific predictive objective}, not as comprehensive clinical narratives. Future work should examine how well the semantic content of these summaries aligns with expert clinical judgment and whether the patterns surfaced by the RL-optimized bottleneck correspond to established behavioral indicators of mental-health change.

\subsection{Limitations and Future Directions}
\label{sub:discussion:limitations}
While \projectname{} demonstrates consistent improvements across benchmarks, backbones, and transfer settings, several limitations should be acknowledged.

First, the framework introduces additional computational and input-scale costs relative to conventional ML baselines. RL-based training samples multiple candidate trajectories per input and performs policy updates over complete two-stage rollouts, which is substantially more expensive than training a gradient-boosted model or supervised objectives, and the two-stage architecture also doubles the number of LLM forward passes at inference time relative to direct prediction. The formulation is further bounded by the LLM's context window: our 14-day windows fit comfortably within current limits, but longer horizons (e.g., months of wearable data or years of electronic health records) or settings with hundreds of concurrent sensor channels would require hierarchical abstraction (summarizing sub-windows and then reasoning over them) or retrieval-augmented selection of the most informative temporal segments before abstraction. These extensions are compatible with the two-stage framework but represent additional engineering effort, and we note that the present gains are achieved with a 4B-parameter model rather than a frontier-scale system.

Second, the quality of the semantic abstraction depends on the LLM's ability to interpret structured behavioral data presented in text form. Our prompt design translates raw sensor features into descriptive labels and interpretable units, but this translation is currently hand-crafted for the behavioral health domain. Applying \projectname{} to new domains requires designing analogous prompt templates that make the domain's temporal data legible to the LLM. While this is a lighter-weight requirement than annotating gold intermediate summaries, it is not fully automatic and introduces a domain-adaptation step that could benefit from further standardization.

Third, our evaluation focuses on ordinal mental-health scores within a narrow range (0--6). The reward function and output format are designed for this setting, and it remains to be validated whether the same approach generalizes to continuous-valued targets, multi-label prediction, or classification tasks with categorical outputs. The core two-stage architecture and RL optimization are agnostic to output format in principle, but the reward design would need to be adapted, and the interaction between reward shaping and summary quality may differ across task types. Exploring how different reward formulations shape the content and structure of the learned abstractions is a promising direction for extending the framework to new domains and task formats.

Fourth, our analysis of the learned semantic abstractions themselves is limited to a few illustrative case study (Sec.~\ref{sub:results:qualitative}). While the cases are consistent with our hypothesis that RL tuning shapes summaries toward decision-relevant behavioral structure, we do not systematically characterize summary content across the full test set. A more complete account would quantify properties such as coverage of behavioral dimensions, temporal localization of claims, lexical diversity, and alignment with clinician-identified patterns, and would examine how these properties correlate with downstream predictive accuracy. Such an analysis would also help clarify when and why summaries fail, for instance, whether errors arise more often from selective compression, from mischaracterization of specific behavioral channels, or from failures of cross-signal integration. This is a natural next step for understanding the mechanism behind \projectname{}'s empirical gains.

%% file: 7-conclusion.tex
\section{Conclusion}
\label{sec:conclusion}
We presented \projectname{}, a two-stage LLM framework for robust longitudinal behavioral time-series modeling that routes prediction through an explicit \emph{semantic bottleneck}, first abstracting raw multivariate sensor trajectories into a natural-language description of behavioral structure and then predicting the downstream outcome from that abstraction alone, with both stages optimized end-to-end via GRPO under RLVR using only validated end-task labels. Instantiated on mental-health prediction from passive sensing, \projectname{} achieved state-of-the-art performance under a rigorous LOSO protocol on a benchmark designed to stress-test cross-cohort generalization, with relative MAE reductions of 3.1--10.1\% and 9.5--44.1\% over strong non-LLM and LLM baselines on anxiety and 3.2--9.6\% and 27.4--57.6\% on depression (all $p$s$<$0.05); these gains held on a second longitudinal sensing benchmark, generalized across four LLM backbones with 38.4--61.6\% relative MAE reductions, and showed promising cross-benchmark transfer ability, while ablations confirmed that neither the two-stage decomposition nor RL alone is sufficient and only their combination produces the observed gains. More broadly, these results suggest that when raw longitudinal sensor data are difficult to generalize across people, contexts, and sensing pipelines, a decision-useful abstraction optimized for a specific predictive objective can serve as a more stable interface between behavioral sensing and downstream inference, pointing to a general direction for RL-tuned LLMs that reason over such semantic abstractions of human behavior for ubiquitous computing.

%% file: 8-appendix.tex
\newpage
\appendix
\section*{APPENDIX}

\section{Feature Sets}
\label{app:feature_sets}
\input{figs/table_globem_feature_set}
\input{figs/table_college_feature_set}

\section{More Qualitative Examples}
\label{app:more_qual_examples}
\input{figs/qualitative_example_3}
\input{figs/qualitative_example_4}

%% file: figs/table_globem_feature_set.tex
\begin{table*}[h]
\caption{
 \textbf{ GLOBEM feature set used in all experiments.} This table details the twelve daily passive-sensing features extracted via the GLOBEM/RAPIDS pipeline, categorized across five behavioral domains. For each 14-day window, the model receives these exact features, alongside their corresponding units and semantic descriptions, to formulate the Stage 1 input.
}
\label{tab:globem_feature_set}
\centering
\scriptsize 
\renewcommand{\arraystretch}{1.0} 
\setlength{\tabcolsep}{3pt}
\begin{tabularx}{\textwidth}{@{}l l l X@{}}
\toprule
\textbf{Domain} & \textbf{Feature key} & \textbf{Unit} & \textbf{Description} \\
\midrule
\multirow{2}{*}{Sleep}
 & \begin{tabular}[t]{@{}l@{}}\texttt{f\_slp:fitbit\_sleep\_intraday\_rapids\_}\\\texttt{sumdurationasleepunifiedmain}\end{tabular} & minutes & Total time asleep during the main sleep period; sleep quantity. \\
 & \begin{tabular}[t]{@{}l@{}}\texttt{f\_slp:fitbit\_sleep\_intraday\_rapids\_}\\\texttt{ratiodurationasleepunifiedwithinmain}\end{tabular} & ratio (0--1) & Proportion of time in bed actually asleep; sleep efficiency. \\
\midrule
\multirow{4}{*}{Mobility}
 & \begin{tabular}[t]{@{}l@{}}\texttt{f\_loc:phone\_locations\_doryab\_}\\\texttt{timeathome}\end{tabular} & minutes & Time spent at the inferred home location; stay-at-home behavior. \\
 & \begin{tabular}[t]{@{}l@{}}\texttt{f\_loc:phone\_locations\_doryab\_}\\\texttt{numberofsignificantplaces}\end{tabular} & count & Number of unique significant locations visited; life-space breadth. \\
 & \begin{tabular}[t]{@{}l@{}}\texttt{f\_loc:phone\_locations\_locmap\_}\\\texttt{duration\_in\_locmap\_greens}\end{tabular} & minutes & Time spent in green spaces or parks; restorative out-of-home context. \\
 & \begin{tabular}[t]{@{}l@{}}\texttt{f\_loc:phone\_locations\_barnett\_}\\\texttt{circdnrtn}\end{tabular} & index (0--1) & Deviation of the daily mobility pattern from the user's norm; routine consistency. \\
\midrule
\multirow{2}{*}{Activity}
 & \begin{tabular}[t]{@{}l@{}}\texttt{f\_steps:fitbit\_steps\_intraday\_rapids\_}\\\texttt{sumsteps}\end{tabular} & count & Total daily step count; overall movement volume. \\
 & \begin{tabular}[t]{@{}l@{}}\texttt{f\_steps:fitbit\_steps\_intraday\_rapids\_}\\\texttt{avgdurationactivebout}\end{tabular} & minutes & Average length of sustained active bouts; intentional activity vs.\ incidental motion. \\
\midrule
\multirow{2}{*}{Phone use}
 & \begin{tabular}[t]{@{}l@{}}\texttt{f\_screen:phone\_screen\_rapids\_}\\\texttt{countepisodeunlock}\end{tabular} & count & Number of phone unlock episodes; device-checking frequency. \\
 & \begin{tabular}[t]{@{}l@{}}\texttt{f\_screen:phone\_screen\_rapids\_}\\\texttt{firstuseafter00unlock}\end{tabular} & minutes & Minutes after midnight until the first unlock; morning-activation timing. \\
\midrule
\multirow{2}{*}{Communication}
 & \begin{tabular}[t]{@{}l@{}}\texttt{f\_call:phone\_calls\_rapids\_}\\\texttt{outgoing\_sumduration}\end{tabular} & minutes & Total outgoing call duration; active social initiative. \\
 & \begin{tabular}[t]{@{}l@{}}\texttt{f\_blue:phone\_bluetooth\_doryab\_}\\\texttt{uniquedevicesothers}\end{tabular} & count & Unique nearby Bluetooth devices; ambient social density. \\
\bottomrule
\end{tabularx}
\end{table*}

%% file: figs/table_college_feature_set.tex
\begin{table*}[h]
\caption{
  \textbf{College Experience feature set used in all experiments.} This table details the fifteen daily features extracted from the College Experience sensing stream, categorized across four behavioral domains. For each 14-day window, the model receives these exact features, alongside their corresponding units and semantic descriptions, to formulate the Stage 1 input.
}
\label{tab:college_feature_set}
\centering
\scriptsize
\renewcommand{\arraystretch}{1.5}
\setlength{\tabcolsep}{3pt}
\begin{tabularx}{\textwidth}{@{}l l l X@{}}
\toprule
\textbf{Domain} & \textbf{Feature key} & \textbf{Unit} & \textbf{Description} \\
\midrule
\multirow{3}{*}{Sleep}
 & \texttt{\scriptsize sleep\_duration} & hours      & Estimated total time asleep for the day; sleep quantity. \\
 & \texttt{\scriptsize sleep\_start}    & 8-min bins & Sleep onset time as offset from 8:00\,PM; bedtime regularity. \\
 & \texttt{\scriptsize sleep\_end}      & 8-min bins & Wake time as offset from 8:00\,PM; morning-activation timing. \\
\midrule
\multirow{8}{*}{Mobility}
 & \texttt{\scriptsize loc\_home\_dur}         & hours  & Time spent at the inferred home location; stay-at-home behavior. \\
 & \texttt{\scriptsize loc\_visit\_num\_ep\_0} & count  & Number of distinct locations visited; life-space breadth. \\
 & \texttt{\scriptsize loc\_dist\_ep\_0}       & meters & Total distance traveled; overall mobility volume. \\
 & \texttt{\scriptsize loc\_social\_dur}       & hours  & Time at social locations; in-person social engagement. \\
 & \texttt{\scriptsize loc\_study\_dur}        & hours  & Time at study locations such as libraries; academic engagement. \\
 & \texttt{\scriptsize loc\_leisure\_dur}      & hours  & Time at leisure locations such as parks or shops; restorative context. \\
 & \texttt{\scriptsize loc\_food\_dur}         & hours  & Time at food locations; routine eating context. \\
 & \texttt{\scriptsize loc\_workout\_dur}      & hours  & Time at workout locations such as gyms; intentional activity. \\
\midrule
\multirow{2}{*}{Activity}
 & \texttt{\scriptsize act\_still\_ep\_0}    & seconds & Total stationary duration; physical inactivity. \\
 & \texttt{\scriptsize act\_walking\_ep\_0}  & seconds & Total walking duration; everyday movement volume. \\
\midrule
\multirow{2}{*}{Phone use}
 & \texttt{\scriptsize unlock\_num\_ep\_0}      & count   & Number of phone unlock events; device-checking frequency. \\
 & \texttt{\scriptsize unlock\_duration\_ep\_0} & seconds & Total time the phone was in the unlocked state; overall device engagement. \\
\bottomrule
\end{tabularx}
\end{table*}

%% file: figs/qualitative_example_3.tex
\begin{figure*}[h]
\centering
\small
\setlength{\tabcolsep}{6pt}
\renewcommand{\arraystretch}{1.12}

\begin{tcolorbox}[
    enhanced,
    colback=white,
    colframe=black!15,
    boxrule=0.5pt,
    arc=2mm,
    left=1.2mm,right=1.2mm,top=1.2mm,bottom=1.2mm
]
\begin{tabularx}{\textwidth}{@{}p{0.18\textwidth}X@{}}
\toprule
\textbf{Example sample} & \texttt{CollegeExperience DS3\_1e85c892\_2019-05-15} \\
\textbf{Gold score} & \textbf{3} \\
\textbf{Predictions} & GPT-5.0: 5 | Vanilla Qwen3-4B: 6 | \textbf{TimeSRL Qwen3-4B: 3} \\
\bottomrule
\end{tabularx}
\end{tcolorbox}

\vspace{0.5em}

\begin{minipage}[t]{0.33\textwidth}
\begin{tcolorbox}[
    enhanced,
    colback=baselinebg,
    colframe=baselinegray!45,
    boxrule=0.8pt,
    arc=2mm,
    left=1.2mm,right=1.2mm,top=1.0mm,bottom=1.0mm,
    title=\textbf{GPT-5.0 (2-Stage baseline)},
    coltitle=black,
    fonttitle=\small\bfseries
]
\textbf{Stage-1 summary excerpt}

\vspace{0.25em}
\footnotesize
Sleep is short and highly irregular, with many nights between $\sim$2--6\,h, a rebound oversleep on D5 (12.2\,h), and an acute low on D14 (1.8\,h). Sleep onset is consistently delayed (often 01:30--05:00), indicating circadian instability. Mobility shows a sustained decline D11--D14 (3.5$\rightarrow$3.0\,h walking) with rising stillness ($\sim$20--21\,h). Phone unlocks peak early (D3--D4: 361, 318) and drop sharply on D13--D14. \textcolor{warnred}{Early in the period, very short sleep co-occurs with elevated phone checking (possible hyperarousal/anxious checking).}

\vspace{0.5em}
\begin{tcolorbox}[
    enhanced,
    colback=warnredbg,
    colframe=warnred!35,
    boxrule=0.5pt,
    arc=1.5mm,
    left=0.8mm,right=0.8mm,top=0.6mm,bottom=0.6mm
]
\footnotesize
\textbf{Compression pattern.} A hyperarousal framing dominates; the late-window unlock drop and mobility decline are read as worsening rather than as part of a mixed picture.
\end{tcolorbox}

\vspace{0.35em}
\textbf{Stage-2 output}

\vspace{0.2em}
\footnotesize
\texttt{<think>} Irregular sleep plus early hyperarousal-style phone checking lean toward moderate-to-severe. \texttt{</think>}

\texttt{score: 5}
\end{tcolorbox}
\end{minipage}\hfill
\begin{minipage}[t]{0.31\textwidth}
\begin{tcolorbox}[
    enhanced,
    colback=baselinebg,
    colframe=baselinegray!45,
    boxrule=0.8pt,
    arc=2mm,
    left=1.2mm,right=1.2mm,top=1.0mm,bottom=1.0mm,
    title=\textbf{Qwen3-4B (2-Stage baseline)},
    coltitle=black,
    fonttitle=\small\bfseries
]
\textbf{Stage-1 summary excerpt}

\vspace{0.25em}
\footnotesize
Significant fluctuations in sleep, with a notable spike on D5 (12.2\,h) followed by a sharp decline to 2\,h on D6. Sleep timing varied widely (D14 ended at 02:40). Phone use stable (1.5--2.5\,h), but unlock counts spiked on D3 (361) and D4 (318). Walking activity 3--7\,h/day with no clear sleep--mobility correlation. Minimal time at external locations (all zeros). \textcolor{warnred}{These patterns suggest anxiety-related sleep disturbances, increased screen reliance, and reduced physical activity, common in anxiety disorders.}

\vspace{0.5em}
\begin{tcolorbox}[
    enhanced,
    colback=warnredbg,
    colframe=warnred!35,
    boxrule=0.5pt,
    arc=1.5mm,
    left=0.8mm,right=0.8mm,top=0.6mm,bottom=0.6mm
]
\footnotesize
\textbf{Compression pattern.} Volatility maps directly to severe anxiety; without a stabilizing-evidence channel, both subscale items are pushed to the maximum.
\end{tcolorbox}

\vspace{0.35em}
\textbf{Stage-2 output}

\vspace{0.2em}
\footnotesize
\texttt{<think>} Significant fluctuations and disruptions; each item $\sim$3, totaling 6. \texttt{</think>}

\texttt{score: 6}
\end{tcolorbox}
\end{minipage}\hfill
\begin{minipage}[t]{0.35\textwidth}
\begin{tcolorbox}[
    enhanced,
    colback=timesrlbluebg,
    colframe=timesrlblue!65,
    boxrule=1.0pt,
    arc=2mm,
    left=1.2mm,right=1.2mm,top=1.0mm,bottom=1.0mm,
    title=\textbf{TimeSRL Qwen3-4B},
    coltitle=black,
    fonttitle=\small\bfseries
]
\textbf{Stage-1 summary excerpt (RL-tuned)}

\vspace{0.25em}
\scriptsize
\textbf{Phone use:} Stable 1.5--2.5\,h baseline with a D4 peak and D7 dip; not the dominant signal.

\textbf{Sleep --- irregular and extreme:} D5 unusually long sleep (12.2\,h), D6 sudden drop to 2\,h, D14 extremely short (1.8\,h). Other days closer to baseline.

\textbf{Activity:} High still time (18--21\,h) with low walking; sedentary throughout.

\textbf{Volatility points:} D5--D6 sharp contrast and D14 acute low --- concentrated, not pervasive.

\textbf{Cross-feature pattern:} \textcolor{goodgreen}{Frequent fluctuations may signal anxiety-induced insomnia or cyclical stressors --- severe enough to flag, but localized to D6 and D14 rather than spanning the whole window.}

\vspace{0.4em}
\emph{Excerpt shown for readability. The full model output also includes additional dimensions such as locations visited and recommendations.}

\vspace{0.5em}
\begin{tcolorbox}[
    enhanced,
    colback=goodgreenbg,
    colframe=goodgreen!35,
    boxrule=0.5pt,
    arc=1.5mm,
    left=0.8mm,right=0.8mm,top=0.6mm,bottom=0.6mm
]
\scriptsize
\textbf{Preservation pattern.} Severe disruptions are flagged as acute and bounded; the rest of the window's relative stability is preserved, supporting a moderate (not maximum) score.
\end{tcolorbox}

\vspace{0.35em}
\textbf{Stage-2 output}

\vspace{0.2em}
\scriptsize
\texttt{<think>} Behavioral data shows clear signs of anxiety, so the score would be 3 --- moderate, not severe. \texttt{</think>}

\texttt{score: 3}
\end{tcolorbox}
\end{minipage}

\vspace{0.4em}

\caption{\textbf{Qualitative comparison on a 14-day CollegeExperience DS3 sample (gold anxiety score = 3).} Model-generated summaries and predictions are presented alongside highlighted framing sentences (colored text) and annotated compression/preservation patterns (shaded boxes). Untuned two-stage baselines misconstrue localized irregularities (e.g., the D5/D6 sleep swing and D14 crash) as a pervasive anxiety pattern, resulting in overestimates (scores of 5 and 6). In contrast, TimeSRL flags these volatile points but localizes them temporally, preserving the relative stability of the remaining window to yield a calibrated moderate-band prediction that matches the ground truth.}
\label{fig:qual_example_ds3_moderate}
\end{figure*}

%% file: figs/qualitative_example_4.tex
\begin{figure*}[h]
\centering
\small
\setlength{\tabcolsep}{6pt}
\renewcommand{\arraystretch}{1.12}

\begin{tcolorbox}[
    enhanced,
    colback=white,
    colframe=black!15,
    boxrule=0.5pt,
    arc=2mm,
    left=1.2mm,right=1.2mm,top=1.2mm,bottom=1.2mm
]
\begin{tabularx}{\textwidth}{@{}p{0.18\textwidth}X@{}}
\toprule
\textbf{Example sample} & \texttt{CollegeExperience DEP DS2\_c4de76cb\_2017-11-13} \\
\textbf{Gold score} & \textbf{2} \\
\textbf{Predictions} & GPT-5.0: 4 | Vanilla Qwen3-4B: 6 | \textbf{TimeSRL Qwen3-4B: 2} \\
\bottomrule
\end{tabularx}
\end{tcolorbox}

\vspace{0.5em}

\begin{minipage}[t]{0.33\textwidth}
\begin{tcolorbox}[
    enhanced,
    colback=baselinebg,
    colframe=baselinegray!45,
    boxrule=0.8pt,
    arc=2mm,
    left=1.2mm,right=1.2mm,top=1.0mm,bottom=1.0mm,
    title=\textbf{GPT-5.0 (2-Stage baseline)},
    coltitle=black,
    fonttitle=\small\bfseries
]
\textbf{Stage-1 summary excerpt}

\vspace{0.25em}
\footnotesize
Across 14 days, a clear drift toward lower daily engagement: walking time and distance traveled decline while stillness increases, locations visited drop, and total phone use / unlocks fall sharply mid-period (nadir D10--D11). Social exposure is near-absent ($\sim$0.5\,h on D7). Sleep is highly irregular --- short-sleep days (D8, D11--12) alternating with hypersomnia (D9--10, D13) and shifting onsets indicating circadian disruption. \textcolor{warnred}{Overall, the trajectory reflects declining movement and phone engagement, persistently minimal social exposure, and unstable sleep --- patterns that can align with low energy and social withdrawal.}

\vspace{0.5em}
\begin{tcolorbox}[
    enhanced,
    colback=warnredbg,
    colframe=warnred!35,
    boxrule=0.5pt,
    arc=1.5mm,
    left=0.8mm,right=0.8mm,top=0.6mm,bottom=0.6mm
]
\footnotesize
\textbf{Compression pattern.} Hedges about exam/illness/sensor confounders are stated but not weighted at scoring time; the global decline-and-withdrawal narrative drives the moderate score.
\end{tcolorbox}

\vspace{0.35em}
\textbf{Stage-2 output}

\vspace{0.2em}
\footnotesize
\texttt{<think>} Trajectory of declining engagement, social withdrawal, and circadian disruption --- weight as moderate. \texttt{</think>}

\texttt{score: 4}
\end{tcolorbox}
\end{minipage}\hfill
\begin{minipage}[t]{0.285\textwidth}
\begin{tcolorbox}[
    enhanced,
    colback=baselinebg,
    colframe=baselinegray!45,
    boxrule=0.8pt,
    arc=2mm,
    left=1.2mm,right=1.2mm,top=1.0mm,bottom=1.0mm,
    title=\textbf{Qwen3-4B (2-Stage baseline)},
    coltitle=black,
    fonttitle=\small\bfseries
]
\textbf{Stage-1 summary excerpt}

\vspace{0.25em}
\footnotesize
Significant irregularities in sleep, social engagement, and physical activity. Sleep fluctuates dramatically: 7.8\,h (D1) → 4.8\,h (D5), spiking to 14.2\,h (D9) and 13\,h (D13). Phone use low ($\le$2.5\,h) with elevated unlocks on D5. Social interaction minimal (zero except 0.5\,h on D7). Notably, only 1.5\,h at home on D11, a stark contrast to the 19--22\,h on other days. \textcolor{warnred}{Over the 14-day period, the individual exhibited significant irregularities in sleep patterns, social engagement, and physical activity, consistent with depressive symptoms.}

\vspace{0.5em}
\begin{tcolorbox}[
    enhanced,
    colback=warnredbg,
    colframe=warnred!35,
    boxrule=0.5pt,
    arc=1.5mm,
    left=0.8mm,right=0.8mm,top=0.6mm,bottom=0.6mm
]
\footnotesize
\textbf{Compression pattern.} Behavioral irregularities are mapped directly to severe depression; both PHQ-4 depression items are pushed to the maximum (3 + 3 = 6).
\end{tcolorbox}

\vspace{0.35em}
\textbf{Stage-2 output}

\vspace{0.2em}
\footnotesize
\texttt{<think>} Each PHQ-4 depression item likely 3, totaling 6. \texttt{</think>}

\texttt{score: 6}
\end{tcolorbox}
\end{minipage}\hfill
\begin{minipage}[t]{0.35\textwidth}
\begin{tcolorbox}[
    enhanced,
    colback=timesrlbluebg,
    colframe=timesrlblue!65,
    boxrule=1.0pt,
    arc=2mm,
    left=1.2mm,right=1.2mm,top=1.0mm,bottom=1.0mm,
    title=\textbf{TimeSRL Qwen3-4B},
    coltitle=black,
    fonttitle=\small\bfseries
]
\textbf{Stage-1 summary excerpt (RL-tuned)}

\vspace{0.25em}
\scriptsize
\textbf{Mobility:} Locations visited dropped from 4--5 (D1--D3) to 1--2 (D6--D10); social-location time near-zero except D7 (0.5\,h).

\textbf{Sleep:} Irregular --- D9 spike (14.2\,h), D8 short ($\sim$3.5\,h); other days mostly 6--8\,h.

\textbf{Activity:} Low walking baseline (0.7--1.4\,h) with a brief D5 peak (4\,h); no other workout activity.

\textbf{Volatility points:} D9 hypersomnia (14.2\,h sleep + 21.2\,h home); D11 reverses to 1.5\,h at home and 1 location visited.

\textbf{Time-localized framing:} \textcolor{goodgreen}{Day 11 is a critical anomaly --- 1.5\,h at home (vs.\ 21+\,h on other days), 1 location, 5.8\,h sleep --- a potential turning point in the behavioral pattern.}

\vspace{0.4em}
\emph{Excerpt shown for readability. The full model output also includes additional dimensions such as cross-feature correlations and recommendations.}

\vspace{0.5em}
\begin{tcolorbox}[
    enhanced,
    colback=goodgreenbg,
    colframe=goodgreen!35,
    boxrule=0.5pt,
    arc=1.5mm,
    left=0.8mm,right=0.8mm,top=0.6mm,bottom=0.6mm
]
\scriptsize
\textbf{Preservation pattern.} D9 and D11 are flagged as discrete volatility points with explicit comparison to surrounding days; the rest of the window remains within typical ranges, keeping the score at the mild-to-moderate boundary.
\end{tcolorbox}

\vspace{0.35em}
\textbf{Stage-2 output}

\vspace{0.2em}
\scriptsize
\texttt{<think>} Symptoms are present but largely concentrated on D9 / D11; weight as the mild-to-moderate boundary rather than severe. \texttt{</think>}

\texttt{score: 2}
\end{tcolorbox}
\end{minipage}

\vspace{0.4em}

\caption{\textbf{Qualitative comparison on a 14-day CollegeExperience DS2 depression sample (gold depression score = 2).} Model-generated summaries and predictions are presented alongside highlighted framing sentences (colored text) and annotated compression/preservation patterns (shaded boxes). The untuned two-stage baselines compress the trajectory into a global decline-and-withdrawal narrative --- GPT-5.0 hedges on confounders (exam/illness/sensor gaps) but still lands at 4, while Qwen3-4B maps the same evidence directly to the maximum (6). TimeSRL flags the D9 hypersomnia day and the D11 anomaly as discrete volatility points and preserves the relative stability of the surrounding window, yielding a calibrated mild-band prediction matching the ground truth.}
\label{fig:qual_example_dep_ds2}
\end{figure*}